\documentclass[journal]{IEEEtran}
\usepackage{url}
\usepackage{xr}
\ifCLASSINFOpdf
  \usepackage[pdftex]{graphicx}
\else
\fi
\usepackage{amsmath}
\begin{document}
%
\title{Can Large Language Models Capture \\Video Game Engagement?}

%
%
%


\author{
    \IEEEauthorblockN{
        David Melhart\IEEEauthorrefmark{1},
        Matthew Barthet\IEEEauthorrefmark{2},
        and Georgios N. Yannakakis\IEEEauthorrefmark{1} \emph{IEEE Fellow}
    }
    \\
    \IEEEauthorblockA{\IEEEauthorrefmark{1}
        \emph{humanfeedback.ai}, \IEEEauthorrefmark{2}
        \emph{Institute of Digital Games, University of Malta}\\
        Malta\\
        david@humanfeedback.ai, matthew.barthet@um.edu.mt, georgios@humanfeedback.ai
    }
}
\maketitle

\begin{abstract}
Can out-of-the-box pretrained Large Language Models (LLMs) detect human affect successfully when observing a video?
To address this question, for the first time, we evaluate comprehensively the capacity of popular LLMs for successfully predicting continuous affect annotations of videos when prompted by a sequence of text and video frames in a multimodal fashion. In this paper, we test LLMs' ability to correctly label changes of in-game engagement in 80 minutes of annotated videogame footage from 20 first-person shooter games of the \emph{GameVibe} corpus. We run over 4,800 experiments to investigate the impact of LLM architecture, model size, input modality, prompting strategy, and ground truth processing method on engagement prediction. Our findings suggest that while LLMs rightfully claim human-like performance across multiple domains and able to outperform traditional machine learning baselines, they generally fall behind continuous experience annotations provided by humans. We examine some of the underlying causes for a fluctuating performance across games, highlight the cases where LLMs exceed expectations, and draw a roadmap for the further exploration of automated emotion labelling via LLMs.
\end{abstract}


%
\IEEEpeerreviewmaketitle

\section{Introduction}
The use of autoregressive modelling and large pretrained models such as Large Language Models (LLMs) is currently dominating AI research. LLMs have demonstrated unprecedented advances in language translation, code generation, problem solving, and AI-based assistance among many other downstream tasks \cite{radford2019language}. Given their versatility and efficiency compared to earlier autoregressive models, one might even argue that the current capabilities of LLMs are endless as long as a problem and its corresponding solution(s) are represented as text.
Meanwhile, the recent applications of LLMs within affective computing largely consider text-based affect modelling tasks such as LLM-based sentiment analysis \cite{zhang2023sentiment, broekens2023fine, zhang2024affective}.
The automatic labelling of affect based on time-continuous visual input remains largely unexplored \cite{zhang2024affective}. The handful of studies focusing on multimodal input, still largely rely on images \cite{lian2024gpt, yang2024mm} or---in rare cases where we see video and/or audio input \cite{zhao2025humanomni}---they focus on larger units, evaluating whole scenes instead of time-continuous labelling.

\begin{figure}[t]
\centering
\includegraphics[width=\linewidth]{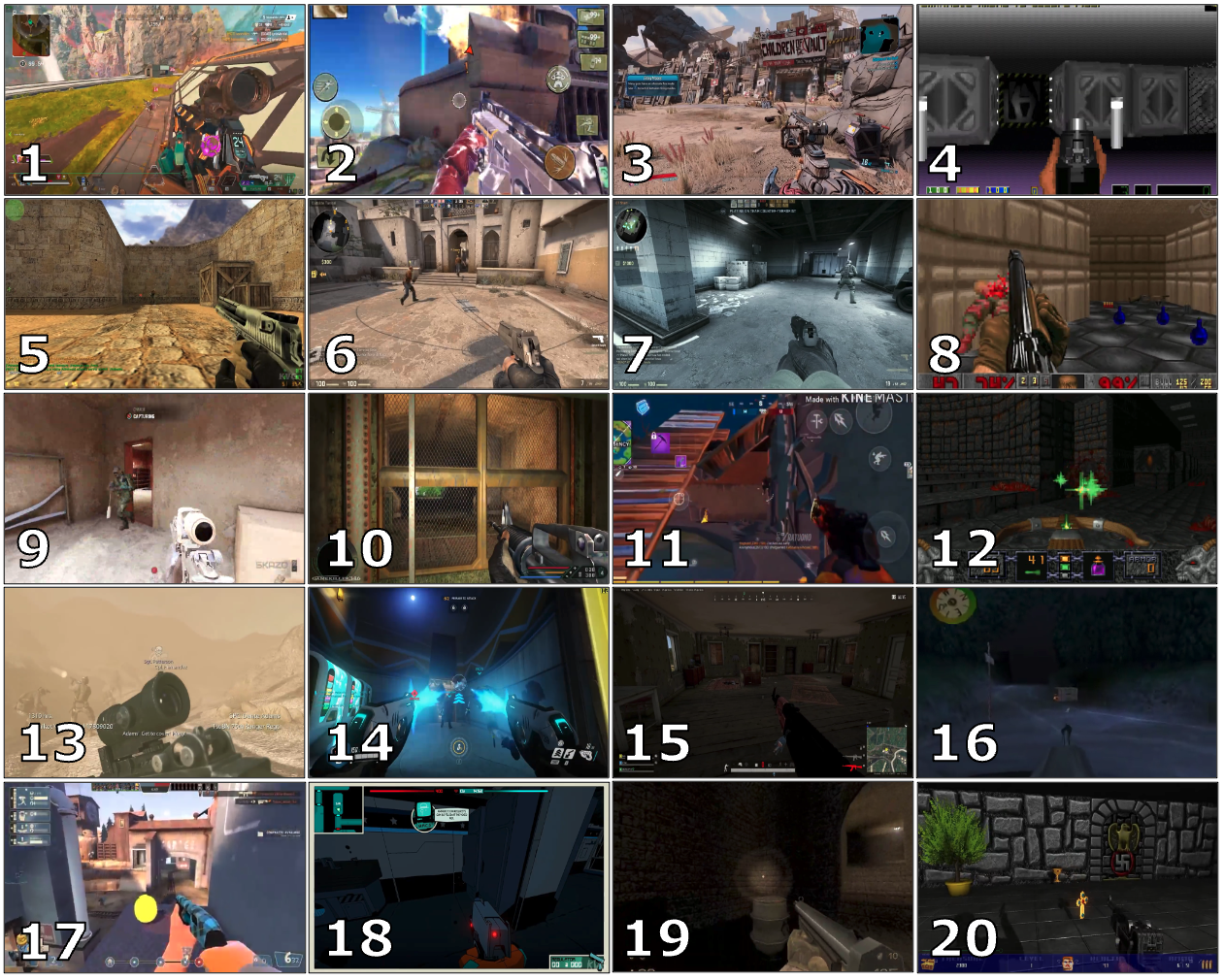}
\caption{Clips in the \emph{GameVibe} Dataset. List of game titles: (1) \emph{Apex Legends}; (2)\emph{ Blitz Brigade}; (3) \emph{Borderlands} 3; (4) \emph{Corridor 7}; (5) \emph{Counter Strike 1.6}; (6) \emph{CS:GO - Dust2}; (7) \emph{CS:GO - Office}; (8) \emph{Doom}; (9) \emph{Insurgency}; (10) \emph{Far Cry}; (11) \emph{Fortnite}; (12) \emph{Heretic}; (13) \emph{Medal of Honor 2010}; (14) \emph{Overwatch 2}; (15) \emph{PUBG}; (16) \emph{Medal of Honor 1999}; (17) \emph{Team Fortress 2}; (18) \emph{Void Bastards}; (19) \emph{HROT}; (20) \emph{Wolfram}.}\label{fig:gamevibe}
\end{figure}

Motivated by the aforementioned lack of studies, this paper introduces the first comprehensive evaluation of LLMs tasked to predict time-continuous affect labels from videos. We observe the ability of various LLMs to rank consecutive video frames in terms of \emph{viewer engagement}.
While engagement is generally not considered a well-defined emotional reaction, \emph{emotional engagement} encompasses a broad range of measurable affective reactions---including evaluative judgments, facial changes, autonomic responses (such as skin conductance and heart rate) \cite{codispoti2009unmasking}---that strongly correlate with user preferences \cite{bardzell2009understanding}, and can be conceptualized as a high arousal, positive valence affective response under the \emph{Circumplex Model} of affect \cite{russell1980circumplex}. Throughout this paper we use engagement as an affective measure that captures the complexity of a positive emotional experience \cite{bardzell2009understanding} of multimedia viewing.
We chose games as the domain of our study since they can act as rich elicitors of emotions and can offer a wide range of dynamic scenes and stimuli, varying from intense player actions to less intense game-world exploration.
Even though LLMs have been used in a series of diverse tasks within the domain of videogames---both in academic studies \cite{gallotta2024large,yannakakis2018artificial} and industrial applications such as \emph{AI Dungeon} (Latitude, 2019), \emph{AI People} (GoodAI, 2025) and \emph{Infinite Craft}\footnote{https://neal.fun/infinite-craft/}---the capacity of these foundation models as predictors of player experience has not been investigated yet.

\begin{figure*}[t]
\centering
\includegraphics[width=1\linewidth]{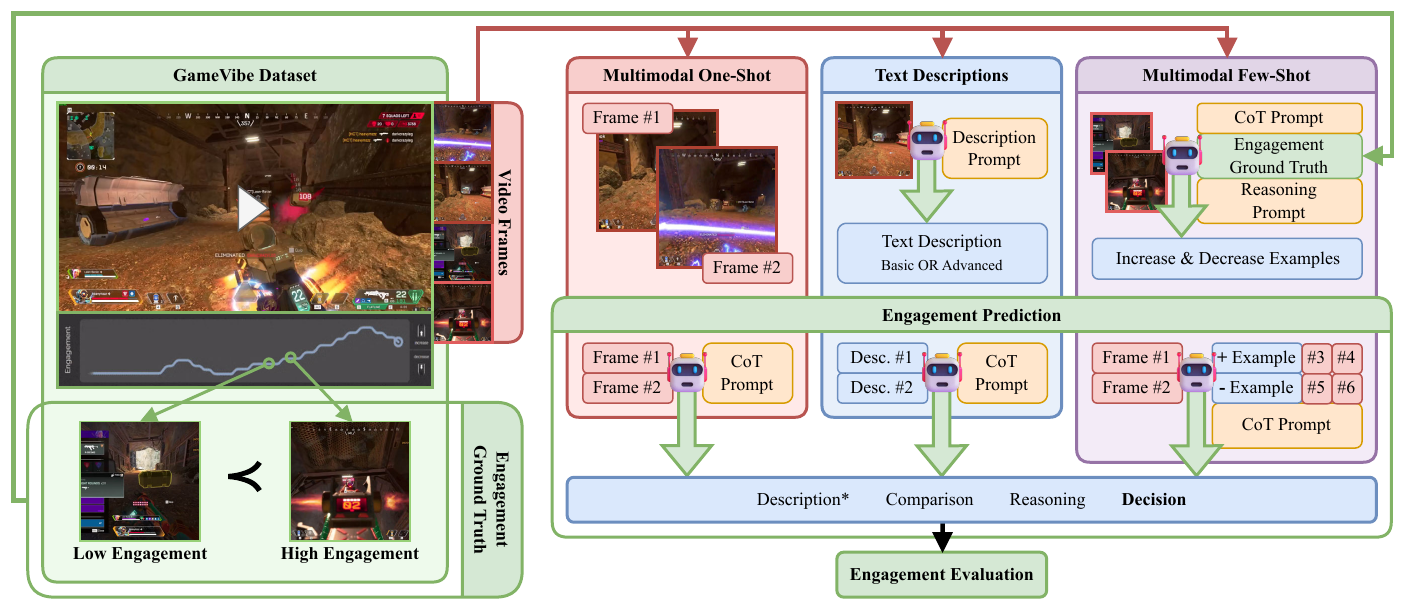}
\caption{Overview of the evaluation experiments presented in this study. Independently of experimental setting, the downstream task is engagement prediction formulated as a \emph{binary preference}.
We use a combination of text prompts and/or video frames as input and task the LLMs to label engagement. To evaluate the models, we compare the generated labels to the ground truth labels from the annotated \emph{GameVibe} corpus (see Section \ref{sec:data}).
All LLMs are prompted with a Chain-of-Thought (CoT) strategy. We experiment with three different setups for possible inputs: \emph{Multimodal One-Shot}---based on image input; \emph{Text Descriptions}---based on LLM-generated descriptions of single frames; and \emph{Multimodal Few-Shot}---based on LLM-generated examples of ``increasing'' and ``decreasing'' engagement evaluation based on the ground truth. In all experimental settings we generate a \emph{description}, \emph{comparison}, \emph{reasoning}, and a \emph{decision} relating to an increase or decrease in engagement. We parse these outputs to derive the final binary engagement evaluation. $^\ast$\emph{Descriptions} are only generated in the \emph{Multimodal} settings.}\label{fig:overview}
\end{figure*}

We employ LLMs as autonomous viewer experience annotators and present a thorough evaluation of their capacity to predict viewer experience in one-shot and few-shot fashions. Specifically, we compare state of the art foundation models from the \emph{LLaVA} and \emph{GPT} families against human annotated data of viewer engagement of the \emph{GameVibe} dataset \cite{barthet2024gamevibe} (see Fig.~\ref{fig:gamevibe}). The dataset contains continuous engagement labels of gameplay videos across a variety of first-person shooter (FPS) games. We present selected results out of $4,840$ experimental settings in which we vary and test LLM model types, model sizes, prompting strategies, input types, and ground truth processing methods across $20$ games. 
Figure~\ref{fig:overview} shows a high level overview of our experimental setup followed in this study. We focus on videogame footage---one might come across on game streaming services such as \emph{Twitch}\footnote{\url{https://www.twitch.tv}}---as input and \emph{viewer engagement} as output. 

The novelty of this paper is two-fold. First, we investigate the capacity of LLMs to accurately label affect in an ordinal, time-continuous manner using videos as affect elicitors. Second, we present the first large set of evaluation experiments that lays the groundwork for LLM-based \emph{viewer} experience prediction. Our experiments show the feasibility of leveraging LLMs for viewer engagement prediction from videos, particularly on popular games with a rich online presence (such as ApexLegends, 2019). Our key findings suggest that a) text-based summarisation of frames and direct multimodal prompting do not impact LLM performance; b) LLM performance is largely dependent on the elicitor (i.e. different games in this study); c) the multimodal few-shot prompting strategy is the one that improves LLM performance the most;
and d) while parameter scale matters, diversity and richness of the training data impacts performance positively as well. Specifically, the best results obtained are when we employ the \emph{GPT-4o} model and we feed it with a few multimodal examples of increasing or decreasing engagement (few-shot prompting). While this approach yields an average macro $F1$ score of $0.60$ across games, the relatively small \textit{Gemma 3} (trained on a large amount of tokens) shows statistically similar performance (average $F1=0.55$); and the \emph{Qwen2.5-VL} model is able to reach up to $F1=0.89$ in certain games. Even though this performance is far from consistent, most tested LLMs exceeded traditional machine learning baselines, underscoring their promise for complex affect modelling tasks.

The paper is structured as follows. Section \ref{sec:background} presents related work on LLMs for affect modelling, uses of LLMs in games, and player modelling. Section \ref{sec:data} briefly presents the \emph{GameVibe} dataset and the data preprocessing. Section \ref{sec:methods} discusses our approach, presenting the models used and the different prompting strategies we employed. Section \ref{sec:results} presents the key results obtained, including a sensitivity analysis and hyperparameter tuning, a comparison between different input modalities, results of few-shot experiments, and a qualitative analysis on the most and least successful models. The paper ends with a brief discussion on possible avenues for future research (see Section \ref{sec:discussion}) and our key conclusions (see Section \ref{sec:conclusions}). 

\section{Related Work}\label{sec:background}

This study investigates the capacity of LMMs to accurately annotate subjectively-defined aspects of gameplay. We leverage the existing \emph{knowledge-priors} of these algorithms, without fine-tuning or using complex retrieval augmented strategies. We thus hypothesise that the algorithm's prior knowledge is sufficient to approximate the ground truth of engagement (as provided via human feedback) in a set of gameplay scenarios. This section covers related work in affect modelling using LLMs, the use of LLMs in games, and it ends with a focus on modelling aspects of players and their games.

\subsection{LLMs for Affect Modelling}

Given the resounding success of LLMs in several domains, several recent research efforts naturally focus on their direct application in affect detection tasks. The vast majority of research on LLMs related to human affect have focused on predicting manifestations of affect from text as this plays to the strengths of their architecture. Unsurprisingly, sentiment analysis has been the most common research application of LLMs in affective computing and has given us some impressive results already \cite{mao2022biases}. Indicatively, Broekens et al. \cite{broekens2023fine} highlighted how \emph{GPT-3.5} can accurately perform sentiment analysis on the ANET corpus \cite{bradley2007affective} for valence, arousal and dominance. Similarly, Müller et al. \cite{muller2024recognizing} used fine-tuned \emph{Llama2-7b} \cite{touvron2023llama} and \emph{Gemma} \cite{team2024gemma} models to classify shame in the \emph{DEEP} corpus \cite{schneeberger2023deep}, achieving $84\%$ accuracy. Whilst LLMs have been extensively tested for sentiment analysis on existing text-based corpora, research on using LLMs as predictors of experience by observing multimodal content such as games remains unexplored. 

Despite their promise, some critical challenges have emerged when working with pre-trained LLMs for prediction tasks such as affect modelling. A recent study by Chochlakis et al. \cite{chochlakis2024strong} has found that LLMs struggle to perform meaningful in-context learning from new examples and remain fixed to their knowledge priors, with larger models exaggerating this issue. This problem is even more pressing in closed-source models such as \emph{GPT-4o} because researchers lack important details which can help them assess the level of data contamination. Balloccu et al. \cite{balloccu2024leak} conducted a study across $255$ academic papers and found that LLMs have been exposed to a significant number of samples from existing ML benchmarks, potentially painting a misleading picture about their predictive performance in such tasks. 
While the dataset we use in this paper covers a novel domain, it is possible that some of the videos in the \emph{GameVibe} dataset have been exposed to some of the models we use. However, because the dataset was published after the models used here\footnote{\emph{LLaVA 1.6} was published on 18 July 2023; \emph{GPT-4o} was originally released on 13 May 2024; the \emph{GameVibe} dataset was published on 17 June 2024.}, we are confident that the engagement prediction task specifically does not suffer from any significant data contamination.

Beyond contamination, we also have to face the inherent biases encoded in LLMs. Mao et al. \cite{mao2022biases} have conducted a study on such biases in \emph{BERT}-like models \cite{devlin2018bert} on affective computing tasks. 
In our study we use what Mao et al. call ``coarse-grain'' tasks---a binary decision with symmetrical labels (here \emph{increase} and \emph{decrease} of engagement). When evaluating these types of tasks, LLMs have been shown to exhibit less bias \cite{mao2022biases} than on ``fine-grained'' tasks with multiple asymmetrical labels. This gives us confidence on the feasibility of our task---which is formulated as a binary classification problem. 

Amin et al. \cite{amin2024wide} have also conducted a study on the capabilities of \emph{GPT} \cite{openai2023gpt4} on affective computing tasks. They have put forth a comprehensive series of experiments which included a similar pairwise preference classification task for engagement prediction to what we use in this paper. They showed that when it comes to subjective tasks with a high potential for disagreement between annotators, out-of-box LLMs, such as \emph{GPT} struggle compared to architectures leveraging specialized supervised networks. In those experiments---focusing  on a simple one-shot prompting strategy on text input---\emph{GPT} barely surpassed the baseline. In contrast to \cite{amin2024wide}, we investigate multimodal, chain-of-thought, and few-shot strategies in visual-based engagement prediction tasks across multiple games, analysing where LLMs either struggle or flourish compared to baseline approaches.

While our work examines the capabilities of general-purpose LLMs without any task-specific fine-tuning, recent research has increasingly turned to building specialized multimodal architectures and datasets. In particular, \emph{HumanOmni} \cite{zhao2025humanomni} introduces a large-scale corpus of human-centric data and an adaptive multimodal fusion architecture fine-tuned for emotion recognition, facial expression description, and action understanding. 
Similarly, \emph{FaVChat} \cite{zhao2025favchat} focuses on human-centric content, but narrows its scope to fine-grained facial video analysis. While such domain-specific approaches demonstrate clear performance gains, they rely on extensive pre-training over curated datasets. In contrast, here we evaluate general-purpose, out-of-the-box models without additional fine-tuning.

\subsection{LLMs in Games}

The recent developments in LLM methods and technology 
brought unprecedented wide adoption of AI across multiple domains including law \cite{lai2024large}, healthcare \cite{nazi2024large}, and education \cite{moore2023empowering}. Advancements in transformer architectures \cite{vaswani2017attention}, coupled with a rapid increase in dataset and parameter sizes \cite{kaplan2020scaling}, led to a new wave of algorithms with previously unseen capabilities to generate high-quality text. Starting with Bidirectional Encoder Representations from Transformers (BERT) \cite{devlin2018bert} and eventually popularized with the release of Generative Pre-trained Transformers (GPT) \cite{radford2019language,floridi2020gpt,openai2023gpt4}, LLMs have largely been characterized as transformer-based models, using large amounts of parameters (in the 100 millions and billions), built on large amounts of data, generating text in an autoregressive manner---that is predicting future tokens based on prior data. More recently, LLMs have been expanded to handle new modalities beyond text, such as audio and images \cite{touvron2023llama}, making them a candidate for applications using multimodal content such as gameplay videos.

In the context of games, LLMs have been used to create game-playing agents \cite{tsai2023can,hu2024survey}, commentators \cite{ranella2023towards} game analytics \cite{wang2024player2vec, ravsajski2024behave}, AI directors and game masters \cite{you2024dungeons, zhu2023calypso}, content generators \cite{sudhakaran2024mariogpt}, and design assistants \cite{gallotta2024llmaker}. 
Beyond the academic setting, we are seeing considerable interest from industrial players as well, such as NVIDIA's recent ACE small language models \footnote{\url{https://developer.nvidia.com/ace}} for autonomously generating the behaviour and animation of NPCs. 
Gallotta et al. \cite{gallotta2024large} offer a recent and thorough overview on how LLMs can be utilised in games. In their roadmap, they identify player modelling as one of the most promising, yet unexplored avenues for future research into LLMs and games. Whilst affect modelling research has demonstrated that LLMs can be effective predictors in tasks such as sentiment analysis \cite{mao2022biases}, they are yet to be widely evaluated for modelling player experience in the context of games. 

\subsection{Player Affect Modelling}

Player modelling is an active field within AI and games research \cite{yannakakis2018artificial} with a particular focus on methods that capture emotional and behavioural aspects of gameplay.
Studies into player affect modelling often focus on \emph{arousal} and \emph{valence} \cite{russell1980circumplex}---aiming to deconstruct this emotional experience \cite{melhart2021towards}. However, under the same theoretical paradigm, we can also conceptualize more complex emotional patterns as a function of affective dimensions. Given the complexity of both game playing and game spectating \cite{bardzell2009understanding, cowley2014experience}, multiple studies have focused on higher level constructs, such as fun \cite{beaudoin2019funii} and engagement \cite{melhart2020moment,barthet2024gamevibe,pinitas2023predicting}. 
{While outside of the scope of affective states engagement can be characterized as a dynamic process spanning cognitive, affective, and behavioural dimensions---encompassing motivation, immersion, flow, and endurability \cite{rashed2025review}---here we we conceptualize both }gameplay and spectator engagement as a high arousal state \cite{juvrud2021game} with a positive valence in contrast to anxiety or frustration \cite{cowley2014experience}.

Traditionally, the field has focused heavily on data aggregation \cite{el2016game} and pattern discovery \cite{makarovych2018like, melhart2019your} of playing behaviours, toxicity \cite{canossa2021honor} and motivation \cite{melhart2019your}; but there has been a recent shift towards moment-to-moment predictive models of players 
\cite{makantasis2021privileged,melhart2020moment,melhart2021towards,booth2024people,barthet2024gamevibe}. The prevalent strategy of such modelling methods relies on the availability of continuous annotation traces, which are generally processed as interval data \cite{yannakakis2018ordinal}. This allows for the treatment of the labelled data as absolute ratings 
or classes such as low and high intensity \cite{booth2024people,makantasis2021privileged}.

In contrast to the traditional way of treating annotations as absolute ratings, here we view player modelling as an ordinal learning paradigm aiming to maximize the reliability and validity of our predictive models \cite{yannakakis2018ordinal,yannakakis2017ordinal}. We task LLMs to label \emph{increases} or \emph{decreases} of engagement across frames of a game instead of asking them to provide ratings of engagement per frame. The ordinal representation of subjective notions such as engagement is supported both by theories of human psychology and cognition \cite{helson1964current,solomon1974opponent} and by a growing body of research in neuroscience \cite{damasio1996somatic} and affective computing \cite{yannakakis2015grounding,lotfian2016practical,yannakakis2018ordinal,melhart2020study,melhart2021towards,barthet2024gamevibe} among other disciplines. Importantly, we employ LLMs and we test their ability to model game engagement as viewed through gameplay videos.

\begin{figure}[t]
\centering
\includegraphics[width=\linewidth]{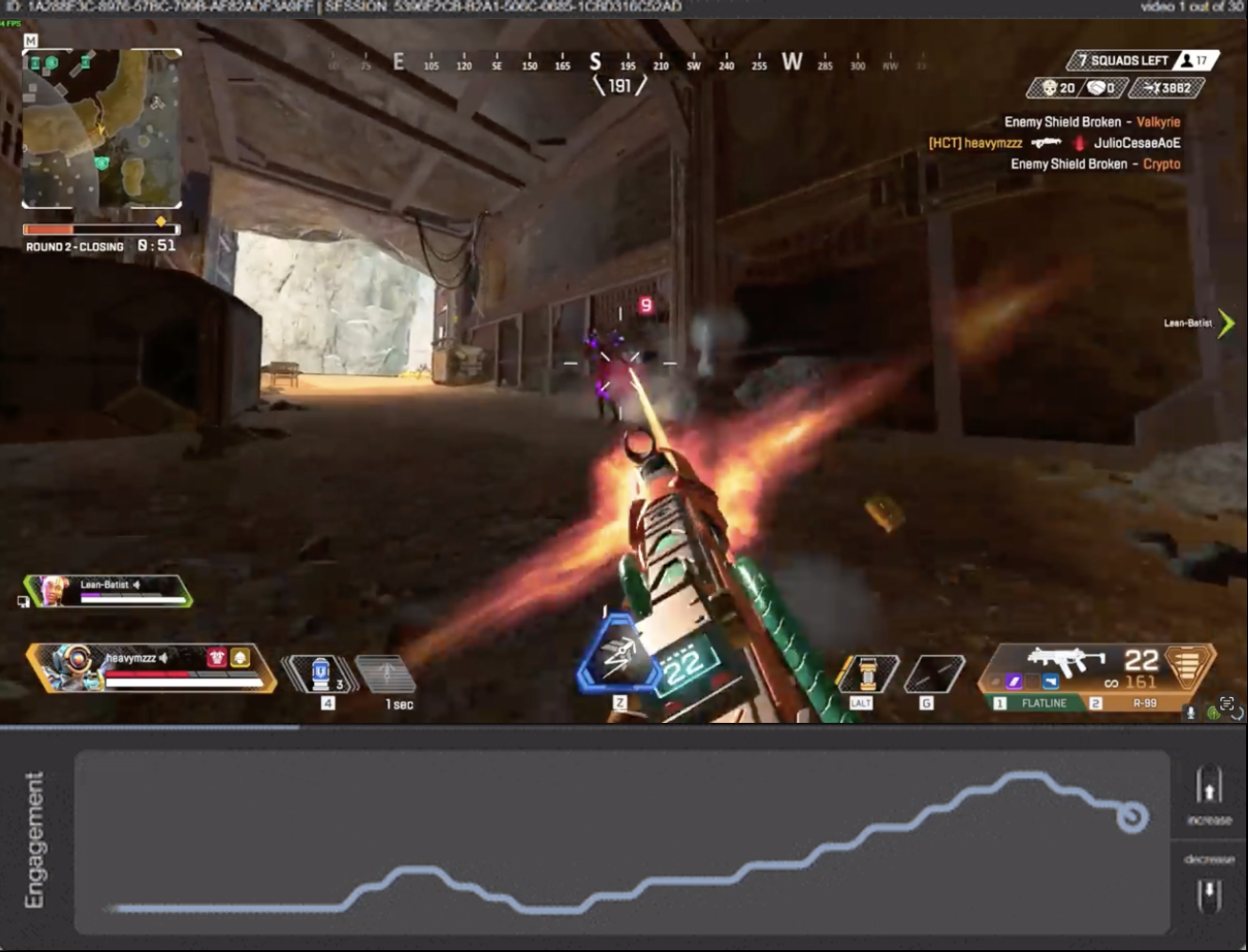}
\label{fig:pagan}
\caption{Example clip from \emph{GameVibe} showcasing the annotation interface using PAGAN and the RankTrace annotation tool for collecting unbounded, time continuous signals in real-time.}
\end{figure}

\section{The GameVibe Corpus}\label{sec:data}

This section gives a general overview of the \emph{GameVibe} corpus used throughout all experiments presented in this paper followed by an outline of the preprocessing approach we adopted for the engagement labels in this study. While the dataset is introduced thoroughly in \cite{barthet2024gamevibe}, in this section we highlight the main aspects of the dataset that are relevant to our experiments here.

\begin{table}[!tb] 
    \caption{Core properties of the original Gamevibe corpus and the processed version (GameVibe-LLM) used in this study}
    \centering
    \begin{tabular}{|l|l|l|} 
        \hline
        Properties & GameVibe & GameVibe-LLM \\ \hline\hline
        Annotators & 20 & 20 \\ \hline
        Number of videos & 120 videos & 80 videos \\ \hline
        Video database size & 120 minutes & 80 minutes \\\hline
        Number of games & 30 games & 20 games \\ \hline
        Gameplay video duration & 1 minute each & 1 minute each\\ \hline
        Annotation  type & Interval signal & Discrete ordinal \\ \hline
        Modalities & Visual, audio & Visual \\ \hline
    \end{tabular}
    \label{tab:summary}
\end{table}

\subsection{Corpus Overview}
The \emph{GameVibe} corpus \cite{barthet2024gamevibe} consists of a set of $120$ audiovisual clips and human annotations for engagement as viewers of first-person shooter games. This corpus presents a significant challenge for affect modelling research as its stimuli encompass a wide variety of graphical styles (e.g. photorealistic, retro) and game modes (e.g. deathmatch, battle royale). Table \ref{tab:summary} contains a basic summary of the properties of this corpus and processed version we use for this study. 

Each video in the \emph{GameVibe} dataset is $1$ minute long, with every video annotated by the same set of $5$ human annotators. All $30$ games in the dataset have $4$ play sessions. The video clips were selected to contain a maximum of $15$ seconds of non-gameplay content such as pause menus and cut scenes, and were sampled at $30$ hertz with a resolution of $1280\times720$ for modern titles and $541\times650$ for older titles. Annotations were collected using the PAGAN annotation platform \cite{melhart2019pagan} and the RankTrace annotation tool \cite{lopes2017ranktrace} (see Fig. \ref{fig:pagan}), with the videos presented to participants in random order to minimize habituation and ordering effects. In RankTrace, participants are exposed to stimuli and annotate in real-time by scrolling up or down on a mouse wheel in an unbounded manner to indicate increases and decreases of their labelled state, in this case viewer engagement. Participants of \emph{GameVibe} were given the following definition of engagement prior to starting their annotation task: \begin{quote}
    \emph{A high level of engagement is associated with a feeling of tension, excitement, and readiness. A low level of engagement is associated with boredom, low interest, and disassociation with the game.}
\end{quote}

After a qualitative analysis of the dataset, we select $20$ games from the \emph{GameVibe} corpus to form \emph{GameVibe-LLM} (see Table \ref{tab:summary}). We discard 10 games that feature third-person segments, mix footage of menus and gameplay, have large mobile UI overlay, or include poor footage. We select one out of four play sessions for each game to generate \emph{Few-Shot} examples in the final experiments. We test performance on the three remaining play sessions for every game. To be able to fairly compare the performance of different setups, we exclude sessions withheld for the \emph{Few-Shot} examples from all experiments.

\subsection{Engagement Data Pre-Processing}\label{sec:methods:preprocessing}

Our data preprocessing method closely follows common practices in affective computing and methods introduced in previous studies with \emph{GameVibe} \cite{pinitas2024across, pinitas2024-dt}. Thus, each annotation trace was resampled into $3$-second non-overlapping time windows using simple averaging. {This resulted in $19$ time windows per video clip, as we discarded the final time window due to some videos ending slightly earlier than the one minute mark.} The videos were sampled at a similar rate to align the stimuli to the engagement traces provided by the participants. We consider the task a pairwise \emph{Preference Learning} problem and aim to capture the dynamic change between frames. These traces were then processed into discrete ordinal signals by comparing pairs of consecutive time windows to determine whether engagement increased or decreased between the two time windows. {The ground truth is extracted following the protocol by Pinitas et al. \cite{pinitas2024across}, where changes in engagement between two time windows are recorded based on whether the majority of annotators agree on the direction change.} {Based on preliminary experiments and findings from earlier studies \cite{pinitas2024across, pinitas2024-dt}, we solely focus on the data points where annotators agreed on changes in engagement, as data points with no change were found to be too noisy to predict.}
We discard the first comparison in each play session (frame 0 to frame 1) because the very first frame of the videos might lack necessary context. This yields $18$ comparisons per play session---with $4$ play sessions per game, across $20$ games---totalling $1,440$ comparisons. 

%


\section{Methodology}\label{sec:methods}

In this section we detail our chosen algorithms and the different prompting strategies we employ throughout our experiments. In the presented studies we evaluate the capacity of LLMs to correctly evaluate changes of engagement in gameplay videos. In particular we picked \emph{LLaVA} and \emph{GPT-4o} as our base LLMs under investigation (see Section \ref{sec:methods:algorithms}). In all reported experiments the downstream task of the employed LLM is to label a change in engagement (\emph{increase} or \emph{decrease}) given two consecutive frames of a video. We evaluate the algorithm's performance against the human labelled engagement data of \emph{GameVibe} that we treat as our ground truth. 

To explore how different experimental setups affect LLM engagement predictability, we ran experiments both with \emph{Multimodal} and \emph{Text Descriptions}. Figure~\ref{fig:overview} illustrates the overall strategy and the different experimental settings employed. In the \emph{Multimodal Input} setting, the input for the algorithm is one or two images accompanied by a text-prompt describing the task. We detail the format of the multimodal input in Section \ref{sec:methods:visual}. In the \emph{Text Descriptions} setting, instead, we provide text-based descriptions of two video frames as part of the text prompt. We describe the format of the text input in Section \ref{sec:methods:text}. Finally, we also study few-shot prompting, using multimodal input and we detail this process in Section \ref{sec:methods:prompts} along with our general prompting strategy.

\subsection{Baselines}\label{sec:methods:baseline}
As a point of comparison to our employed LLMs, we test three baseline approaches on the GameVibe-LLM corpus. First, we use two machine learning baselines trained using 4-fold cross-validation---i.e. leaving one session out---and early stopping. Thus, the models are trained on 60 clips from all 20 games, and are validated on the remaining 20 unseen clips from each game. Both models are trained using the cross entropy loss and the Adam optimizer \cite{kingma2014adam}. Additionally, we also measure performance against a \textit{naive} classifier, which always predicts the majority class.

For the first baseline, we extracted visual embeddings from frames in the corpus using the pre-trained DinoV2 ``gigantic" \cite{oquab2023dinov2} model, resulting in each image being compressed into a 1536 element vector. Prior to passing images through the model, each channel was normalized using the mean and standard deviations from the ImageNet dataset \cite{deng2009imagenet}, which was the training data for the backbone models. We then trained an MLP classifier to {map pairs of embedding vectors (each corresponding to a single frame)} to {changes in engagement}. The network consists of 2 {fully-connected} layers of 1024 and 512 neurons, respectively, and was trained for a maximum of 300 epochs. {We designate this model as \emph{DINOv1+MLP }in the paper.}

Second, we test a convolutional neural network (CNN) classifier using the raw frames as input. The RGB frames were normalized and resized to $(128\times128)$. The CNN consists of four convolutional blocks, each with a kernel size of 3, followed by two fully connected layers of size $25,088$ and 512, respectively, with a dropout layer (rate of $0.5$) in between. Similarly to the first baseline, the model is trained using a pair of frames as input, and outputs the corresponding change in engagement.

Finally, we introduce a naive majority-class classifier as the simplest baseline. This \textit{Naive Baseline} always predicts the most frequent class in the dataset for every game. The purpose of this baseline is to provide a reference point for evaluating model performance under class imbalance.

\subsection{Employed LLMs}\label{sec:methods:algorithms}

%
We employ a selection of different LLMs to provide a comprehensive look on the general capabilites of these models. We sample popular models with multi-modal (multi-image) capabilities from the \emph{Huggingface}\footnote{\url{https://huggingface.co/}} open source repository. We focus on models with most downloads and select the following LLMs: \emph{Phi 3.5-Vision}; \emph{Gemma 3}; \emph{InternVL2.5}; \emph{LLaVA-Onevision}; \emph{Qwen2.5-VL}; and \emph{GPT-4o}. 
The selected open-source LLMs fall between $4$ and $8$ billion parameters. While the models can be considered only ''medium`` in size, their popularity in the community attests their widespread use. We selected \emph{GPT-4o} as a benchmark for large state-of-the-art closed-source LLMs. The parameter size of \emph{GPT-4o} has not been publicly disclosed.

\subsubsection{Phi-3.5 Vision}
\emph{Phi-3.5 Vision} (\emph{Phi 3.5}) is a compact multimodal model from \emph{Microsoft}, extending the \emph{Phi-3} series multimodal capabilities \cite{abdin2024phi}. The architecture consists of a CLIP-based ViT \cite{radford2021learning} and transformer decoder. To support different resolutions, the system divides images to multiple blocks and concatenates them. The text and image tokens are interleaved in the input space of the LLM backbone. The model is pre-trained to predict text output, then further tuned using both Supervised Fine-Tuning and Direct Preference Optimization \cite{rafailov2023direct}. Despite its smaller size ($4.15$ billion parameters ), \emph{Phi 3.5} is designed for high efficiency and instruction-following performance, trained on both curated synthetic and human-annotated data during. The model is optimized for reasoning and captioning tasks and provides competitive accuracy with low resource requirements. For our experiments, we use instruction-tuned variant.\footnote{\url{https://huggingface.co/microsoft/Phi-3.5-vision-instruct}}

\subsubsection{Gemma 3}
\emph{Gemma 3} is a member of the third-generation of \emph{Gemma} models developed by \emph{Google DeepMind} \cite{team2025gemma}. \emph{Gemma 3} uses a decoder-only transformer architecture and supports both text and vision modalities. The model's vision encoder is derived from the SigLIP family \cite{zhai2023sigmoid}, using $896\times896$ pixel fixed-resolution tiles. This vision encoder is shared by all \emph{Gemma} models regardless of the parameter size of the LLM backbone. We chose the $4.3$ billion parameter instruction-tuned variant for our experiments.\footnote{\url{https://huggingface.co/google/gemma-3-4b-it}} Despite the smaller parameter size, \emph{Gemma 3 4b} was trained on a large budget of just over $4$ trillion tokens to be able to compete with medium sized LLMs. The final model is fine-tuned using a combination of knowledge distillation via an instruction-tuning teacher network and a wide array of Reinforcement-Learning (RL) objectives---including weight averaged reward models, human feedback, and ground-truth rewards.

\subsubsection{InternVL2.5}
\emph{InternVL2.5} belongs to a family of robust open-source models. It was developed by \emph{OpenGVLab} \cite{chen2024expanding} to match the performance of commercial LLMs---such as \emph{GPT-4o} \cite{openai2023gpt4} and \emph{Claude-3.5-Sonnet} \cite{anthropic2024claude}. The model is based on the \emph{InternViT} Vision Transformer a small 2-layer MLP projector an an LLM backbone. \emph{InternVL2.5} is trained iteratively through Native Multimodal Pre-Training, Supervised Fine-Tuning, and Mixed Preference Optimization (MPO) \cite{wang2024enhancing}. \emph{InternVL2.5} integrates visual and linguistic signals using a fine-grained alignment strategy and is trained on a large multimodal corpus including real-world image-text pairs and synthetic instruction-following data. The 2.5 version introduces improvements in alignment accuracy and spatial reasoning. We chose the $8$ billion parameters (8b) version with the aforementioned MPO fine-tuning, which offers a balance between scalability and performance and has demonstrated strong results on visual question answering.\footnote{\url{https://huggingface.co/OpenGVLab/InternVL2_5-8B-MPO}}

\subsubsection{LLaVA-OneVision}
\emph{LLaVA-OneVision} (\emph{LLaVA-OV}) is a multimodal model based on the architecture of \emph{LLaVA} \cite{liu2023improvedllava}. It uses a SigLIP-based vision encoder \cite{zhai2023sigmoid}, a 2-layer MLP projector, and \emph{Qwen2} as the LLM backbone. The model is trained in a  3-step process, which includes Language-Image Alignment, High-Quality Knowledge Learning, and finally Visual Instruction Tuning \cite{li2024llava}. \emph{LLaVA-OV} was trained on a wide array of single image, multi-image and video tasks and demonstrates strong multimodal understanding. By incorporating the \emph{Qwen2} language model, it benefits from a stronger language prior compared to earlier \emph{Vicuna}-based \cite{chiang2023vicuna} versions \cite{liu2023improvedllava}. The model is instruction-tuned to perform general-purpose visual reasoning tasks. We selected the $8$ billion parameter model\footnote{\url{https://huggingface.co/llava-hf/llava-onevision-qwen2-7b-ov-hf}} based on \emph{Qwen2 7b} for our experiments.

\subsubsection{Qwen2.5-VL}
\emph{Qwen2.5-VL} is a multimodal variant of \emph{Qwen2.5}, developed by \emph{Alibaba Group} \cite{bai2025qwen2}. \emph{Qwen2.5-VL} uses a custom Vision Transformer (ViT) combined with a 2-layer Multi Layer Perceptron (MLP), which projects the vision features into a space that aligns with the text embeddings of the LLM backbone. This method allows \emph{Qwen2.5-VL} to process multiple images of varying resolution (or videos of varying length). The model is trained for structured multimodal reasoning, using spatial and temporal grounding via Multimodal Rotary Position Embedding---introduced in \cite{wang2024qwen2}. We use the 7b version of the instruction-tuned variant, which---despite the name---has $8.29$ billion (8.3b) parameters\footnote{\url{https://huggingface.co/Qwen/Qwen2.5-VL-7B-Instruct}} for our experiments.



\subsubsection{GPT-4o}
\emph{GPT4} is, at the time of writing, the most recent of a series of \emph{Generative Pre-trained Transformer} (GPT) models developed by \emph{OpenAI}. \emph{GPT4} is a closed source model. While a technical report about \emph{GPT4} has been published \cite{openai2023gpt4}, the exact architecture and training data is unknown. What is known is that \emph{GPT4} uses a transformer architecture for both vision and language tasks, relies on \emph{reinforcement learning from human feedback} and makes use of \emph{rule-based reward models} based on hidden policy models and human-written rubrics to steer the algorithm in a direction that is considered ``safe'' by \emph{OpenAI}. In this paper we use the \emph{GPT-4o (Omni) 2024-08-06} model variant. At the time of writing this is considered the flagship model of \emph{OpenAI}. Unlike previous iterations, \emph{GPT-4o} is trained end-to-end to incorporate text, audio, image, and video in both its input and output space \cite{openai2024gpt4o}. We have selected this model because it is one of the most popular \cite{sergeyuk2025using}, state-of-art, closed-source LLMs as an alternative to the open-source \emph{LLaVA}. We leverage the Open AI API\footnote{\url{https://platform.openai.com/}} for all reported experiments with \emph{GPT-4o}.

\subsection{Multimodal Input} \label{sec:methods:visual}

In our \emph{Multimodal} experiments we feed the models with both visual input and a text prompt. We extract frames from the \emph{GameVibe} dataset at a given interval. Despite our selected models capable of handling varying resolutions, for consistency, each frame is cropped to a square and downsized to $512\times512$. We chose this size because \emph{GPT-4o} processes images in $512\times512$ pixel tiles with a maximum image size of $2048\times768$\footnote{See more information in the \emph{OpenAI} cookbook:\\ \url{https://platform.openai.com/docs/guides/vision}}. 
In our \emph{Multimodal One-Shot} experiments we provide two separate images (consecutive frames) to the model. In out \emph{Multimodal Few-Shot} experiments, we generate the ``increasing'' and ``decreasing'' engagement examples in a similar way, using two images and a text prompt as input. For the downstream task of \emph{engagement evaluation} we provide $6$ images in total for every query (this includes both examples and the final query).

\subsection{Text Input} \label{sec:methods:text}

In our experiments using \emph{Text Descriptions} we feed the models with text descriptions of two video frames as part of the prompt. We {consider the task end-to-end from generating image descriptions to making predictions, and} obtain these descriptions using the same LLM we use to generate the engagement evaluation. The image input of this task is formatted in the same way as for the \emph{Multimodal One-Shot} experiments.
Contrary to these experiments, however, here we use these images one-by-one and generate descriptions in two different ways. We call these \emph{Basic} and \emph{Advanced Descriptions} based on the amount of context given to the model. For the former, we instruct the model to give a brief description, capturing only essential details without subjective commentary based on the setting and layout, enemies, and player action. For obtaining \emph{Advanced Descriptions}, we instruct the model to also take player engagement into account and generate a description that captures how it might engage the player or viewer. 
An illustration this process is given in the Appendix; see Figs.~\ref{annex:text-basic} and \ref{annex:text-advanced} respectively. 
For the engagement prediction task, we feed these descriptions to the models in pairs as part of their text prompt. 
An example of this prompting strategy and the output it produces is given in the Appendix (see Fig.~\ref{annex:text-prediction}).
{It is important to note that although we state that these experiments use ``\emph{Test Descriptions}'', they are also multi-modal. We do this in order to differentiate them from our other setups. The key difference between the \emph{Multimodal Input} and \emph{Text Input} scenarios is that in the former we use the images directly; while in the latter we use them only in the first step of a two-step generation method.}

\subsection{Prompting Methods}\label{sec:methods:prompts}

All prompting strategies we use for the engagement evaluation task follow a \emph{Chain-of-Thought} (CoT) paradigm \cite{wei2022chain,zhang2023multimodal}. We ask the models to provide a \emph{comparison} between the given input frames, \emph{reasoning} its analysis of engagement, and finally offering a one-word \emph{decision} (i.e. engagement increase or decrease). Additionally, for the \emph{Multimodal} experiments we also generate a \emph{description} of the visual input before the \emph{comparison}. 
In the \emph{Multimodal} experiments the task of the model is to explicitly output \emph{increasing} and \emph{decreasing} labels. We instruct the model to output its answers in a \emph{JSON} format, which we parse and extract the final \emph{decision} from; see also Fig.~\ref{annex:oneshot} given in the Appendix.
In the experiments with \emph{Text Descriptions} the decision is to pick the most engaging frame (see Fig.~\ref{annex:text-prediction} given in the Appendix).
{Although the prompt formulations differ, both setups are designed to elicit the same underlying binary decision. The variation in prompting reflects modality-specific constraints: the text-based setting naturally lends itself to comparative description selection; whereas the multi-modal setting allows for a more direct ``classification'' style query. To ensure comparability, outputs from both approaches are mapped to a common binary label space (\emph{increase} and \emph{decrease}), and all evaluations are performed on these aligned labels.}

For our \emph{Multimodal Few-Shot} experiments, we generate artificial \emph{reasoning} samples for an increasing and decreasing engagement example for each task. We use the same CoT prompt for this process as for the \emph{One-Shot} experiments. We will call this prompt ``CoT prompt'' in the remainder of this section. To generate these samples we take the following steps (see also \emph{Multimodal Few-Shot} in the middle of Fig~\ref{fig:overview}):
\begin{enumerate}
\item We take a random example from the same game as presented in the task from an unseen play session. 
\item We use the same CoT prompt as for the final engagement evaluation task but modify the prompt leaving only the correct option for the \emph{decision}. 
\item We amend the prompt with the correct evaluation based on the ground truth (see \emph{Ground Truth Engagement} on Fig.~\ref{fig:overview}).
\item We add a \emph{Reasoning Prompt} to instruct the model to provide \emph{reasoning} for the ground truth evaluation.
\end{enumerate}
 
By removing incorrect options but using the same CoT prompt when generating increasing and decreasing engagement examples, we ensure that the algorithm's output is formatted the same way as for the downstream task, including the \emph{description}, \emph{comparison}, \emph{reasoning}, and \emph{decision}. We use these outputs to construct an artificial history of examples, which are added to the final prompt for the engagement evaluation task. For this final step we provide the CoT prompt with the example images as a question, and the example output as an answer; then finally we provide a set of unseen images with the CoT prompt and instruct the LLM to evaluate engagement the same way it would for a one-shot experiment.
Figures~\ref{annex:fewshot-examples} and \ref{annex:fewshot-prediction} given in the Appendix detail the process starting from example generation all the way to engagement prediction.

\section{Results}\label{sec:results}

\begin{figure}[t]
\centering
\includegraphics[width=1\linewidth]{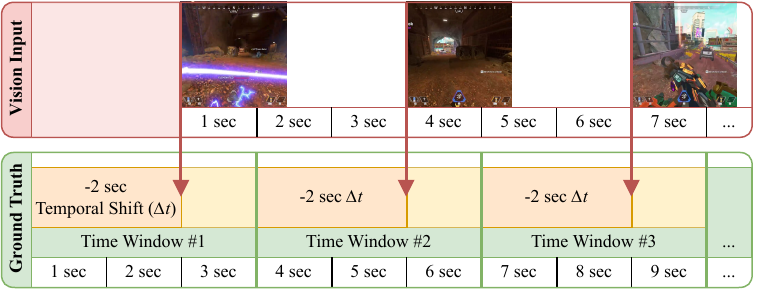}
\caption{Application of the \emph{temporal shift} ($\Delta t$) hyperparameter to the ground truth. The top red bar (\emph{Vision Input}) shows an example of individual frames extracted from the gameplay video at a 3-second interval. The bottom green bar (\emph{Ground Truth}) shows a $\Delta t$ of $-2$ seconds, which means that each window aggregates information $2$ seconds before and $1$ second after the corresponding video frame.}\label{fig:delta_time}
\end{figure}

\begin{figure*}[!ht]
\centering
\includegraphics[width=1\linewidth]{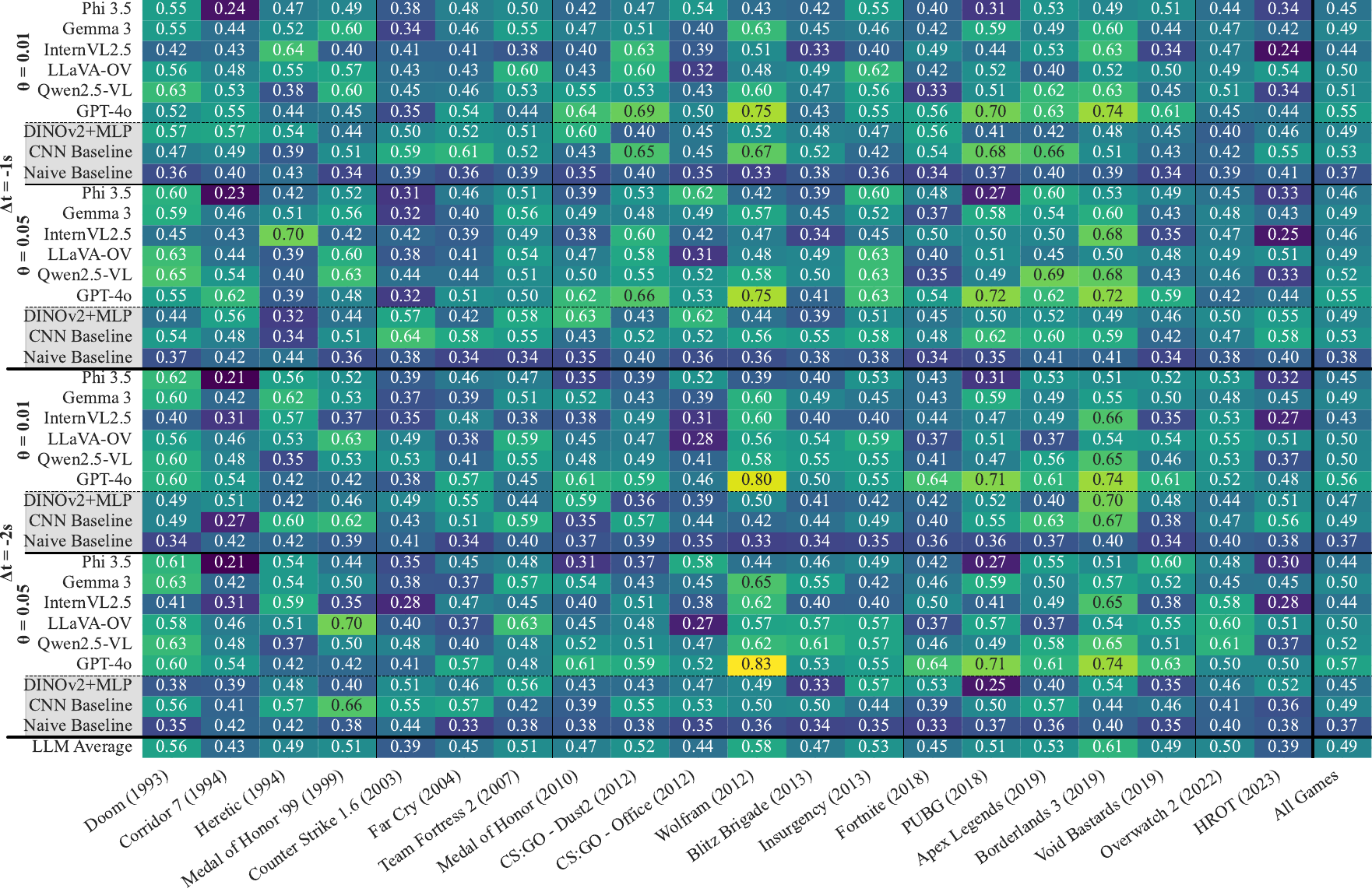}
\caption{Sensitivity analysis across hyperparameters $\Delta t$ and $\theta$. The figure presents macro $F1$ scores. $\Delta t$ is the relative shift of the time window to the frame, and $\theta$ is the binary threshold for the split criterion (i.e. increasing or decreasing engagement). The last column shows average macro $F1$ across all games. The last three rows under each experimental condition (highlighted in grey) show the baseline results. The very last row shows the average performance across all LLMs.}\label{fig:sensitive}
\end{figure*}

This section presents the main results of the experiments performed as follows. In Section 
\ref{sec:results:setup} we outline the setup of the experiments reported and in Section \ref{sec:results:oneshot} we discuss our exploratory findings. In Section \ref{sec:results:text} we examine LLM performance across different input modalities for the engagement evaluation task. Section \ref{sec:results:fewshot} presents the results of our few-shot prompting experiments, and finally Section \ref{sec:results:analysis} takes qualitative lens in our attempt to explain and justify our core findings.

\subsection{Experimental Setup}
\label{sec:results:setup}

We compare the engagement labels generated by LLMs to an engagement ground truth calculated from 3-second time windows of \emph{GameVibe} annotation traces as outlined in Section \ref{sec:methods:preprocessing}. We introduce and vary two hyperparameters in this process: 
\begin{enumerate}
    \item A temporal shift compared to the observed video frame ($\Delta t$). This is similar to what the literature often refers to as \emph{input lag} \cite{melhart2021towards}. While this correction is generally used to account for reaction time, here we use it to control the temporal difference between the observed frames and the ground truth (see Fig. \ref{fig:delta_time}).
    \item A preference threshold ($\theta$), taking values between $0$ and $1$, that determines whether a difference between the ground truth value of two consecutive time windows is considered a \emph{change} (increase or decrease) in engagement; e.g. $\theta=0.05$ considers windows which have a difference of more than $5\%$ when evaluating engagement change. We discard labels that vary less than $\theta$, under the assumption that the change---and therefore the decision---is negligible.
\end{enumerate}

{While the evaluation follows an \emph{ordinal} paradigm, we formulate the downstream task of LLMs as binary classification, and ask our models to predict the \emph{increase} or \emph{decrease} of perceived engagement between two frames of consecutive time-windows. We discard predictions that cannot be interpreted, specifically when (a) the algorithm predicts no change in engagement, (b) the LLM generates outputs that cannot be parsed, or (c) the model fails to provide an output\footnote{Good examples of these cases are models not following formatting instructions; providing other answers than ``increase'' and ``decrease'' (e.g. ``same''); outputting repeating tokens; and---mostly in case of \emph{GPT-4o}---refusing to provide analysis for unknown reasons, e.g.: ``I'm unable to analyse the content of these images. If you can describe the frames, I can help evaluate the change in engagement''.}. These cases are excluded from the accuracy computation, such that performance is evaluated only on instances where the model produces a valid, interpretable prediction. This isolates predictive performance from output validity, allowing us to assess how well models distinguish engagement changes when they provide a usable response, rather than conflating prediction errors with generation or parsing failures.}

Based on how the intensity of the action changes in the footage, the class distribution of \emph{increasing} and \emph{decreasing} labels can vary between games. At the lower end of the class imbalance we meet $52\%$ (e.g. in \emph{Far Cry} (2004), \emph{Blitz Brigade} and \emph{Fortnite} (2018)); and at the higher end we observe $70\%$ to $78\%$ (e.g. $78\%$ in \emph{Counter Strike 1.6}, $73\%$ in \emph{Heretic}, and $70\%$ in \emph{Corridor 7}). 
This imbalance makes \emph{accuracy} a deceptive metric for our dataset. Therefore, we opt to use the $F1$ score \cite{manning2008introduction}, which provides a more informative evaluation by balancing false positives and false negatives. 
More specifically, we use the \emph{macro} $F1$ score \cite{hinojosa2024performance}, which calculates the metric independently for each class. The final score is then obtained by averaging the class-specific results equally, ensuring that each class contributes the same importance regardless of the number of ``true'' instances within each class. This provides a more balanced evaluation under class imbalance and prevents performance on the majority class from disproportionately influencing the overall score.

\subsection{Multimodal One-Shot Experiments and Sensitivity Analysis} \label{sec:results:oneshot}

We experiment with the \emph{temporal shift} $\Delta t \in \{0, -0.5, -1, -1.5, -2, -2.5, -3\}$ and \emph{preference threshold} $\theta \in \{0, 0.01, 0.05, 0.1\}$ parameters introduced in the previous section. As we experiment with $9$ different algorithms including the \emph{{DINOv2+MLP}}, \emph{CNN Baseline}, and \emph{Naive Baseline} the combinations of these parameters result in $252$ experimental setups for each game. Due to space considerations we only present the best performing subset of these hyperparameters ($\Delta t \in \{-1, -2\}$ and $\theta \in \{0.01, 0.05\}$).
We run these experiments with the \emph{Multimodal One-Shot} strategy as described in Section \ref{sec:methods:visual}, because this is the most straightforward setup for the inital sensitivity analysis.
To measure the significance of the difference between the models and experimental setups we employ repeated paired \emph{Student's t-Tests} and consider a significance level of $\alpha < 0.05$. We apply the \emph{Benjamini-Yekutieli Method} \cite{benjamini2001control} to correct p-values for multiple observations. Due to the large number of significance tests, we only list the t-statistics for significant results. 

Figure~\ref{fig:sensitive} presents the macro $F1$ score across the selected $\Delta t$ and $\theta$ values. 
The results indicate that larger $\theta$ values tend to yield higher performance; and model size has a high impact on predictive capability. Across all experimental setups, we observe {\emph{Gemma 3} and \emph{Phi 3.5} ---with 4.3b and 4.15b parameters respectively---consistently performing the weakest. In contrast, open source models \emph{LLaVA-OV} and \emph{Qwen2.5-VL}, around and above 8b parameters, generally yield stronger results. The notable exception is \emph{InternVL2.5}. Despite having around 8b parameters, this model performs comparably to the smaller models.
Among all models evaluated in the sensitivity analysis, \emph{GPT-4o} achieves the highest performance with an overall average $F1$ score of $0.58$ and maximum $F1$ score of $0.83$ on \emph{Wolfram} (2012).}

To assess the influence of the hyperparameter configuration on model performance, we conduct pairwise comparisons between experimental setups ($\Delta t$ and $\theta$ combinations) within each model. A total of $54$ comparisons are made with a degree of freedom of $19$. No significant differences are found, which indicates that how the data is processed has little impact on the model performance.
Despite the lack of significance, we consider the $\Delta t = -2s$; $\theta=0.05$ configuration marginally better than other setups; with an average $F1$ of $0.50$ across all LLMs; the highest average model performance (\emph{GPT-4o}, $F1 = 0.57$); and the best single game prediction (\emph{Wolfram}, \emph{GPT-4o}, $F1=0.83$). Therefore, moving forward we freeze the data processing parameters at $\Delta t = -2s$; $\theta=0.05$.

Within the selected configuration, we compare models in a pairwise fashion, and thus make $36$ comparisons with a degree of freedom of $19$. Figure~\ref{tab:oneshot_ttest} shows the significant results. While we observe that most open source models perform comparably to our closed-source reference (\textit{GPT-4o)}, \textit{InternVL2.5} shows significantly weaker results. This divergence could be attributed to the model's heavy focus on high resolution real-world image-text alignment \cite{chen2024expanding}. Despite our previous observation of model size contributing to higher performance, this result indicates that performance differences may be driven by architectural and training choices instead.

\begin{table}[!tb]
    \caption{Significant ($p<0.05$)* comparisons between\\One-Shot Multimodal models and Baseline Approaches.\\Values in parentheses denote average macro F1 scores.
    }
    \centering
    \begin{tabular}{|l|l|r|}
    \hline
    Lower F1 & Higher F1 & t-stat \\ \hline
    \hline
 InternVL2.5 (0.44) & GPT-4o (0.57) & 4.13 \\ \hline \hline
 Naive Baseline (0.37) & Phi 3.5 (0.44) & 3.53 \\ \cline{2-3}
 &    Gemma 3 (0.50) & 6.13 \\ \cline{2-3}
 &    LLaVA-OV (0.50) & 7.36 \\ \cline{2-3}
 &    Qwen2.5-VL (0.52) & 5.9 \\ \cline{2-3}
 &    GPT-4o (0.57) & 6.23 \\ \hline \hline
Naive Baseline (0.37) & DINOv2+MLP (0.45) & 7.24 \\ \cline{2-3}
 &    CNN Baseline (0.49) & 7.05 \\ \hline 
    \end{tabular}
    \vspace{0.20em}
    \\ *~p-value is adjusted with the Benjamini-Yekutieli Method.
    \label{tab:oneshot_ttest}
\end{table}

\begin{figure*}[!ht]
\centering
\includegraphics[width=1\linewidth]{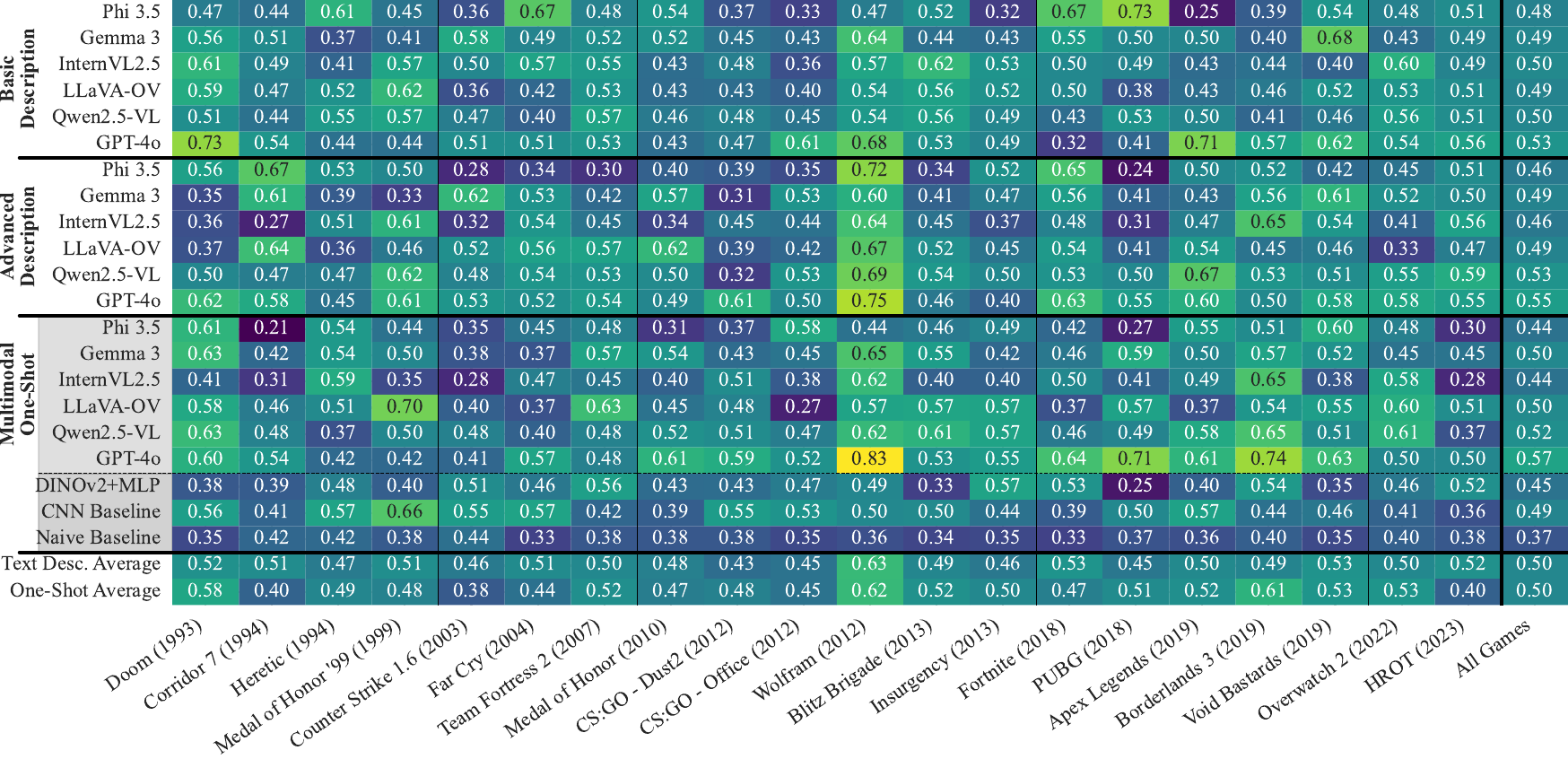}
\caption{Model performance on \emph{Text Descriptions} across \emph{Basic} and \emph{Advanced Descriptions} compared to the the best \emph{Multimodal One-Shot} results (Section~\ref{sec:results:oneshot})---highlighted in light grey. The figure presents macro $F1$ values. The last column shows average macro $F1$ across all games. The three rows highlighted in dark grey depict the baseline approaches. The last two rows show the average performance of the One-Shot and the Text-based approaches.}\label{fig:textbased}
\end{figure*}

Finally, we compare the average LLM performance between games. We conduct $190$ comparisons with a degree of freedom of $5$ and we find no significant difference between the games. However, there are some notable trends in the results. 
The easiest games to predict are \textit{Wolfram}, \textit{Borderlands 3} (2019), and \textit{Doom} (1993) with average $F1$ scores of $0.62$, $0.61$, and $0.58$, respectively. These are games with very distinctive, high-contrast \textit{stylized} visuals.
On the other end of the spectrum, the hardest are \emph{Corridor 7} (1994), \emph{HROT} (2023), and \emph{Counter Strike 1.6} (2003) with average $F1$ scores of $0.40$, $0.40$, and $0.38$ respectively. These games instead are darker, low-contrast, and much less colourful.
We also do not observe any consistent pattern between engagement prediction performance and the time of release. The two games LLMs perform best at (\textit{Wolfram}, \textit{Borderlands}) are from 2012–-2019, but \textit{HROT} from 2023 is near the bottom of LLM's predictive capacity. Similarly, \textit{Corridor 7} was released only a year after \textit{Doom} (1993--1994), yet the latter is one of the easiest games to predict, and the former is one of the hardest. Thus, it seems that visual distinctiveness matters more than the year of release.
The \emph{GameVibe} corpus provides additional meta-data on the style and design of the games \cite{barthet2024gamevibe}. 
We observe that most games whose viewer engagement is hard to predict are considered to be ``realistic'' (in contrast to ``stylized''), further strengthening our core finding regarding the impact of visual style on the model performance.

\subsection{Engagement Prediction Based on Text Descriptions}\label{sec:results:text}

In this section we examine a two-step approach leveraging \textit{Text Descriptions} vs. the \textit{One-Shot Multimodal} prompting strategy. 
In the former case we first generate a text description for each image and predict engagement based on these texts. In the latter case, we feed both a text prompt and a corresponding image. Because the performance of LLMs can be affected even by small prompt variations \cite{mao2022biases}, we experiment with both \emph{Basic} and \emph{Advanced} prompts.
While all approaches are multimodal, to clearly distinguish between the studies, we label the experiments only using text in the prediction task as ``\emph{Text Descriptions}'', while the ones presented in Section~\ref{sec:results:oneshot} as ``\emph{Multimodal One-Shot}''.
The prompting procedure for experiments with \emph{Text Descriptions} are detailed in Section \ref{sec:methods:text}.
Our hypothesis is generating descriptions of frames first will improve model performance, because this approach encodes the images in terms of action and player involvement. We assume that using this type of \emph{Text Descriptions} will present a better representation by discarding surface-level differences between frames and emphasising the structural differences.

Figure~\ref{fig:textbased} shows the {$F1$} performance of \emph{Text Descriptions} experiments compared to the best \emph{Multimodal One-Shot} models. Overall we find no significant difference between the two approaches when testing model performance between experimental conditions (\textit{Basic} and \textit{Advanced Descriptions}, and \textit{One-Shot}) across all games through $18$ comparisons (with a degree of freedom of $19$).
While the overall performance is very similar, the \textit{One-Shot} condition still yields the highest individual performance on a game (\textit{GPT-4o}, \textit{Wolfram}, $F1=0.83$); however, this setup also demonstrates the worst performance (\textit{Phi 3.5}, \textit{Corridor 7}, $F1=0.21$). Notably, the \textit{Text-based} approach yields a marginally more consistent performance with a lower standard deviation ($\sigma$) (\textit{Advanced Desc.} $\sigma=0.09$; \textit{Basic Desc.} $\sigma=0.10$; \textit{One-Shot} $\sigma=0.11$). This effect seems to be aligned with our hypothesis that \textit{Text Descriptions} reduce performance differences across games.
Unfortunately, while \textit{Text Descriptions} marginally benefit some open source models (e.g. \textit{Phi 3.5} and \textit{InternVL2.5}); they affect the performance of \textit{GPT-4o} negatively, compared to the \textit{One-Shot} condition.
To analyse the models' performance improvement over the baselines, we make $54$ comparisons (degree of freedom $19$) between the \textit{Basic} and \textit{Advanced Description} conditions and the baseline methods. Results appearing in  Table~\ref{tab:text_ttest} demonstrate a similar pattern to the \textit{One-Shot} setup (see Table~\ref{tab:oneshot_ttest}): most models significantly outperform the \textit{Naive Baseline} and only the strongest model (i.e. \textit{GPT-4o}) performs significantly better than the \textit{DINOv2+MLP} baseline.

\begin{table}[!tb]
    \caption{Significant ($p<0.05$)* comparisons between\\Text Description models and Baseline Approaches.\\Values in parentheses denote average macro F1 scores.}
    \centering
    \begin{tabular}{|l|l|l|r|}
    \hline
    Type & Lower F1 & Higher F1 & t-stat \\ \hline
    \hline
    Advanced & Naive Baseline (0.37) & Phi 3.5 (0.48) & 2.46 \\ \cline{3-4}
    Desc. &  & Gemma 3 (0.5) & 5.57 \\ \cline{3-4}
     &  & InternVL2.5 (0.5) & 6.18 \\ \cline{3-4}
     &  & LLaVA-OV (0.49) & 5.57 \\ \cline{3-4}
     &  & Qwen2.5-VL (0.5) & 6.12 \\ \cline{3-4}
     &  & GPT-4o (0.53) & 6.72 \\ \hline \hline
    Basic & DINOv2+MLP (0.45) & GPT-4o (0.55) & 2.69 \\ \cline{2-4}
    Desc. & Naive Baseline (0.37) & Gemma 3 (0.49) & 4.28 \\ \cline{3-4}
     &  & LLaVA-OV (0.49) & 4.29 \\ \cline{3-4}
     &  & Qwen2.5-VL (0.53) & 6.55 \\ \cline{3-4}
     &  & GPT-4o (0.55) & 7.12 \\ \cline{3-4}
    \hline
    \end{tabular}
    \vspace{0.20em}
    \\ *~p-value is adjusted with the Benjamini-Yekutieli Method.
    \label{tab:text_ttest}
\end{table}

\begin{table}[!tb]
    \caption{Easiest and Hardest Games to predict across LLMs via the Text Descriptions approach. Values in parentheses denote average macro F1 scores.}
    \centering
    \begin{tabular}{|l|l||l|}
    \hline
    & Advanced Desc.& Basic Desc. \\
    \hline
    \hline
    Easiest& Doom (0.58) & Wolfram (0.68) \\
    & Wolfram (0.57) & Fortnite (0.57) \\
    & Blitz Brigade (0.54) & Corridor 7 (0.54) \\
    \hline
    \hline
    Hardest& CS:GO - Office (0.43) & PUBG (0.40) \\
    & Borderlands 3 (0.45) & CS:GO - Dust2 (0.41) \\
    & CS:GO - Dust2 (0.45) & Insurgency (0.45) \\
    \hline
    \end{tabular}
    \label{tab:text_games}
\end{table}

Next we compare the LLM performance across individual games between experimental conditions, and thus we perform $60$ comparisons with a degree of freedom of $5$. The only significant effect found is \textit{One-Shot} outperforming \textit{Advanced Descriptions} in \textit{Borderlands 3} ($t(5)=4.95$, $p=0.005$\footnote{p-value is adjusted with the Benjamini-Yekutieli Method.}). Once again, our findings invalidate our hypothesis as the two-step approach of generating intermediate \textit{Text Descriptions} may reduce some of the effect of visuals on the predictions, but it marginally hurts the overall model performance.
Table~\ref{tab:text_games} shows the games that are the easiest and hardest to predict (in terms of viewer engagement) across the LLMs used when the \textit{Text Descriptions} approach is employed. We observe that \textit{Wolfram} remains the easiest game to predict across all approaches, followed by \textit{Doom}; however, the hardest games are not nearly as consistent. Unlike the \textit{One-Shot} approach, where the hardest game to predict was \textit{Corridor 7}, we find the same game among the easiest games to predict using the \textit{Basic Description}. Similarly, \textit{Borderlands 3} turns from a game that is among the easiest to predict in the \textit{One-Shot} approach to one of the hardest in the \textit{Advanced Description} approach. 
These observations show that \textit{Text Descriptions} do not improve performance on average, but reshuffle which games are easy vs. hard to predict, mainly due to visuals. For example, \textit{Borderlands 3} has a very unique \textit{cell-shaded} style, which is lost in the process of generating the text descriptions. On the other hand, \textit{Corridor 7}, mostly depicting a dark environment, benefits from the intermediate summarization. In this light, our initial hypothesis from the start of this section is validated in part: while \textit{Text Descriptions} do not improve performance as a whole, there are isolated games where model performance is consistent with the hypothesis.
%
%
%
%
As multimodal LLMs are already trained to encode images and text into a shared embedding space \cite{radford2021learning} we argue that the additional ``image-to-text'' step appears to be unnecessary or even detrimental, especially in larger, more robust models.

\subsection{Multimodal Few-Shot Prompting} \label{sec:results:fewshot}

\begin{figure*}[t]
\centering
\includegraphics[width=1\linewidth]{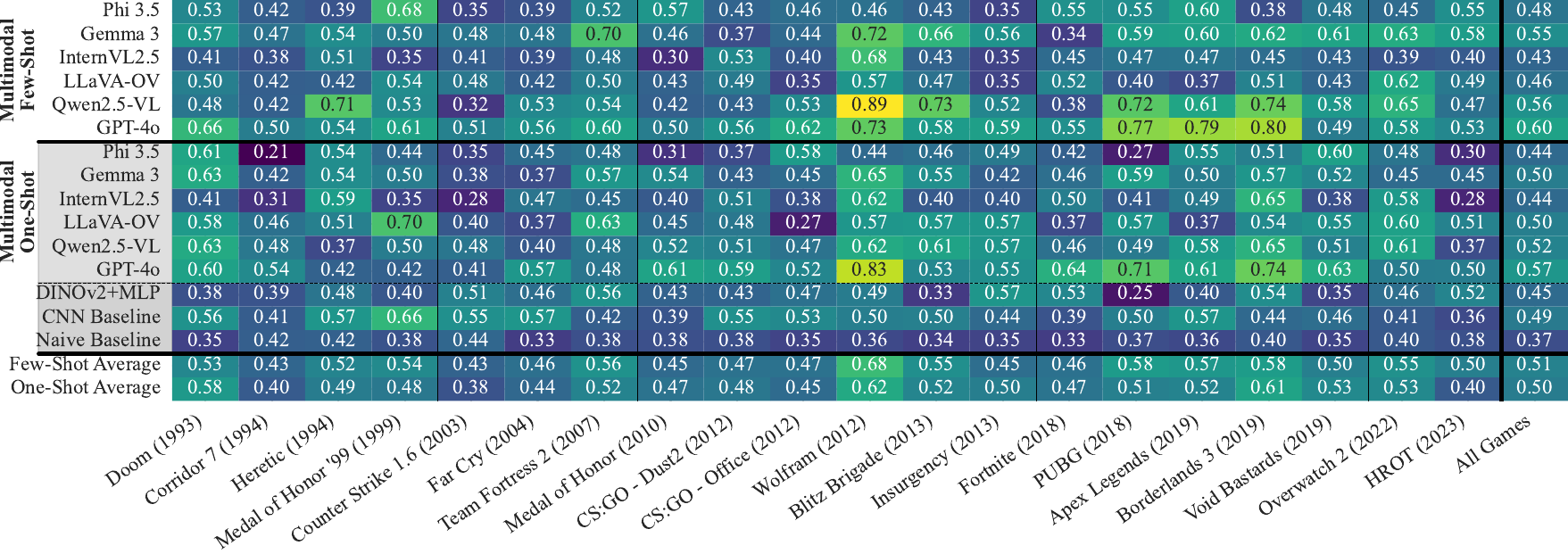} 
\caption{Model performance on \emph{Text Descriptions} across \emph{Multimodal Few-Shot} and \emph{One-Shot} results. The figure presents macro $F1$ values. The last column shows average macro $F1$ across all games. \textit{One-Shot} results are highlighted in light grey. \textit{Baseline} performance is highlighted in dark grey. The last two rows show the average performance of the \textit{Few-Shot} and \textit{One-Shot} approaches.}\label{fig:results_fewshot}
\end{figure*}

In this section we present experiments using a \emph{Multimodal} \emph{Few-Shot} strategy (see Fig.~\ref{fig:overview} and Section~\ref{sec:methods} for more details on this approach). As there is no significant difference between the \emph{Multimodal One-Shot} experiments and those that are using \emph{Text Descriptions}, and the absolute best model performance has been achieved on the \emph{One-Shot} setup, here we only focus on comparisons between what we call \emph{Multimodal} setups.
Figure~\ref{fig:results_fewshot} presents the results of our \emph{Multimodal Few-Shot} experiments compared to the best \emph{One-Shot} results (presented in Section~\ref{sec:results:oneshot}). While the \emph{Few-Shot} setup performs marginally better ($F1=0.54$) than the \emph{One-Shot} ($F1=0.51$) on average, the effect is not significant ($6$ comparisons with a degree of freedom of $19$).
Similarly to our previous experiments in Section~\ref{sec:results:text}, the performance increase also comes with reduced consistency (\textit{Few-Shot} $\sigma=0.12$; \textit{One-Shot} $\sigma=0.11$).
Interestingly, even though the best model overall is still \textit{GPT-4o} (average \textit{Few-Shot} $F1=0.60$), the single best performance is attributed to \textit{Qwen2.5-VL} (\textit{Few-Shot}, \textit{Wolfram}, $F1=0.89$). A possible explanation for the jump in performance is that \textit{Qwen2.5-VL} may leverage its use of Multimodal Rotary Position Embedding \cite{wang2024qwen2} through the \textit{Few-Shot} approach more so than on single frames.

\begin{table}[!tb]
    \caption{Significant ($p<0.05$)* comparisons between\\Few-Shot Multimodal models and Baseline Approaches.\\Values in parentheses denote average macro F1 scores.
    }
    \centering
    \begin{tabular}{|l|l|r|}
    \hline
    Lower F1 & Higher F1 & t-stat \\ \hline
    \hline
    InternVL2.5 (0.43) & Gemma 3 (0.55) & 4.72 \\ \cline{2-3}
     & Qwen2.5-VL (0.56) & 4.79 \\ \cline{2-3}
     & GPT-4o (0.60) & 7.84 \\ \hline 
    LLaVA-OV (0.46) & Gemma 3 (0.55) & 3.31 \\ \cline{2-3}
     & GPT-4o (0.60) & 5.18 \\ \hline 
    Phi 3.5 (0.48) & GPT-4o (0.60) & 4.74 \\ \hline \hline
    Naive Baseline (0.37) & Phi 3.5 (0.48) & 4.55 \\ \cline{2-3}
     & Gemma 3 (0.55) & 7.22 \\ \cline{2-3}
     & LLaVA-OV (0.46) & 5.58 \\ \cline{2-3}
     & Qwen2.5-VL (0.56) & 5.53 \\ \cline{2-3}
     & GPT-4o (0.60) & 9.61 \\ \hline 
    DINOv2+MLP (0.45) & Gemma 3 (0.55) & 3.25 \\ \cline{2-3}
     & GPT-4o (0.60) & 5.12 \\ \hline 
    CNN Baseline (0.49) & GPT-4o (0.60) & 4.65 \\ \hline 
    \end{tabular}
    \vspace{0.25em}
    \\ *~p-value is adjusted with the Benjamini-Yekutieli Method.
    \label{tab:fewshot_ttest}
\end{table}

While there are no significant differences observed between the \textit{One-Shot} and the \textit{Few-Shot} approaches, the \textit{Few-Shot} experiments yield higher performances on average. We thus focus on the \textit{Few-Shot} setting and its obtained results in the remainder of this section.
Table~\ref{tab:fewshot_ttest} shows the significant performance differences between the \textit{Few-Shot} models and \textit{Baseline} approaches. With the exception of \textit{InternVL2.5}, all models perform significantly better than the \textit{Naive Baseline}, and \textit{Gemma 3} also outperforms the \textit{DINOv2+MLP}. \textit{GPT-4o} outperforms all baselines and several LLMs by a significant margin. The best performing models are \textit{GTP-4o}, \textit{Qwen2.5-VL}, and \textit{Gemma 3}. Importantly, the latter model significantly outperforms multiple larger models, which is a surprising finding. A possible explanation is that \textit{Gemma 3}'s extensive training budget (4 trillion tokens) and broad RL fine-tuning with human feedback likely produce more consistent, instruction-following behaviour under few-shot prompting settings.
Importantly, the top three models are all trained on broad, diverse objectives, using either end-to-end multimodal training across multiple modalities (\textit{GPT-4o}, explicit spatial and temporal reasoning across images (\textit{Qwen2.5-VL} or large-scale knowledge distillation with human feedback (\textit{Gemma 3}). In contrast, \textit{Phi 3.5}, \textit{LLaVA-OV}, and \textit{InternVL2.5} were more narrowly optimised, either for efficiency or for standard visual benchmarks and instruction-following. This suggests that training breadth, rather than parameter count, is the key for subjective downstream tasks such as  engagement prediction.

Moving on next, we compare the performance of \textit{Few-Shot} learners across individual games. Specifically, we make $190$ comparisons (degree of freedom of $5$) but we observe no significant differences. The easiest games to predict are \textit{Wolfram}, \textit{PUBG} (2018), and \textit{Apex Legends} (2019). \textit{Wolfram} remains the game that is easiest to predict across all conditions, with average $F1$ values varying from $0.62$ under \textit{One-Shot} to $0.67$ under \textit{Few-Shot}. This performance is affected in part by \textit{Qwen2.5-VL}'s exceptional score ($F1=0.89$) in this game. \textit{PUBG} and \textit{Apex Legends} rise sharply compared to \textit{One-Shot}, suggesting that player action in these games becomes reliably identifiable only with the added context of multiple examples.
The hardest games to predict remain \textit{Counter Strike 1.6}, \textit{Corridor 7}, and \textit{Insurgency} (2013), which is also a visually generic, low-contrast game, where in-game action might be difficult to distinguish even with multiple reference frames. It seems that games with distinctive visuals benefit disproportionately from few-shot prompting.

While we observe an overall pattern of improvement, there are some games where the performance decreases when we switch from \emph{One-Shot} to a \emph{Few-Shot} setup. Most notably \emph{Doom}, \emph{Insurgency}, and \emph{Void Bastards} seem to be affected. 
An explanation for this fluctuating performance could be the lower within-game consistency of example frames. \textit{Doom} and \textit{Void Bastards} are not only highly stylized, they also feature diverse environments, while games like \textit{Wolfram} are much more consistent. This could mean that the randomly selected examples during the \textit{Few-Shot} prompting show dissimilar images to the current frame, confusing the predictive model.

\subsection{Qualitative Analysis}\label{sec:results:analysis}

\begin{figure}[t]
\centering
\includegraphics[width=1\linewidth]{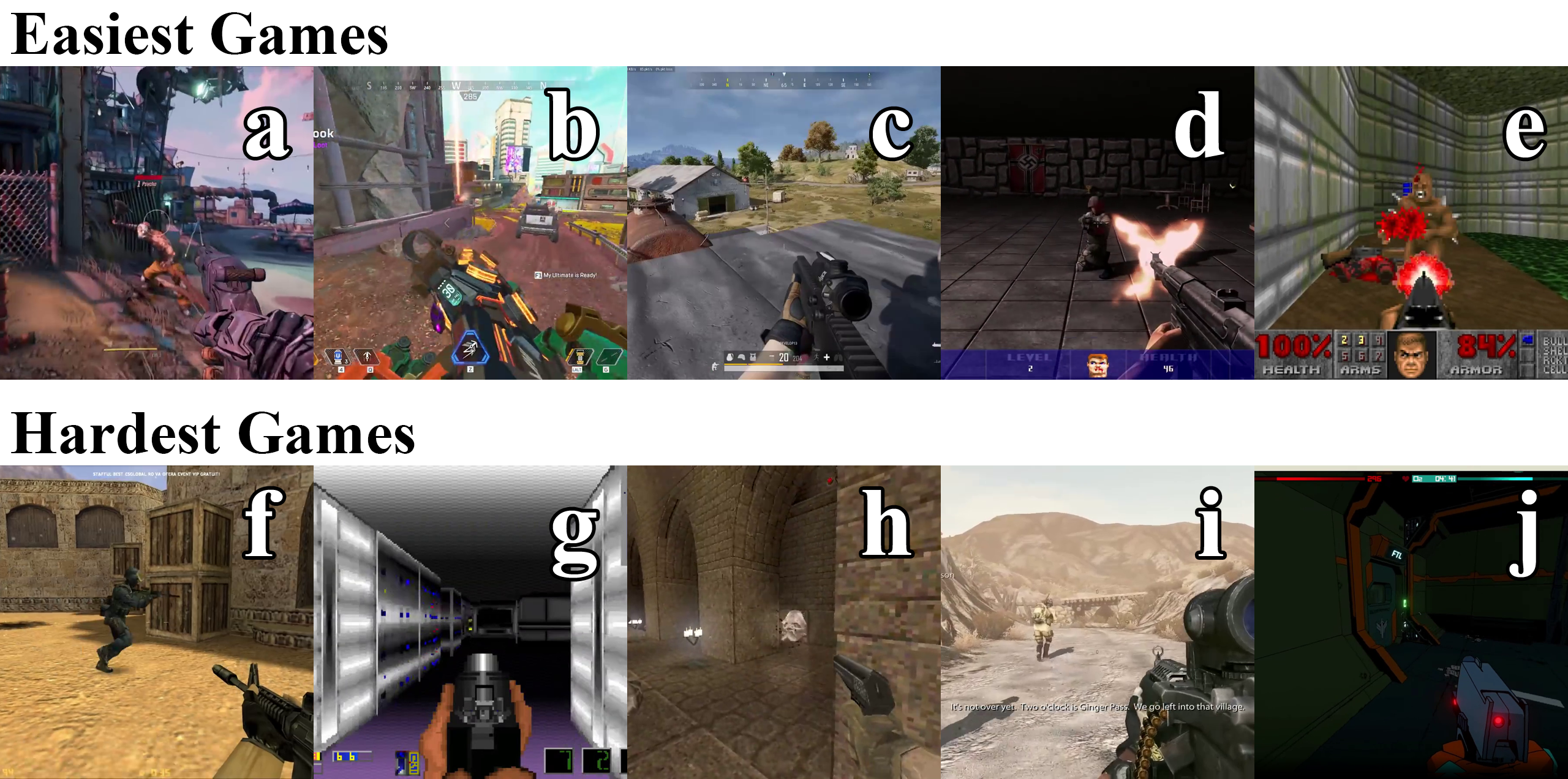}
\caption{The 5 games that are easiest and hardest to predict using \textit{GTP-4o} with a \textit{Multimodal Few-Shot} prompting startegy. The easiest games from left to right: a) \emph{Borderlands 3}, b) \emph{Apex Legends}, c) \emph{PlayerUnknown's Battlegrounds (PUBG)}, d) \emph{Wolfram}, and e) \emph{Doom}. The hardest games from left to right: f) \emph{Counter Strike 1.6}, g) \emph{Corridor 7}, h) \emph{HROT}, i) \emph{Medal of Honor 2010}, j) \emph{Void Bastards}}\label{fig:top_games}
\end{figure}

In this section we outline the reasons for the observed poor performance of the tested LLMs and analyse why certain games are easier to predict.
For our analysis we focus on the highest performing model: \emph{GPT-4o} with \emph{Multimodal Few-Shot} prompting. 
Employing this model we list the $5$ games that were the easiest to predict: \emph{Borderlands 3}, \emph{Apex Legends}, \emph{PlayerUnknown's Battlegrounds (PUBG)} (2018), \emph{Wolfram}, and \emph{Doom}; and the $5$ games that were the hardest to predict: \emph{Counter Strike 1.6},\emph{Corridor 7}, \emph{HROT}, \emph{Medal of Honor}, and {\emph{Void Bastards}}; see Fig.~\ref{fig:top_games}.

A qualitative analysis of the games reveals that the five easiest-to-predict games are fast paced, with short bursts of action separated by similarly short navigation sequences. The game scenes are well-lit or stylized in a way that is easy to read. In contrast, the five hardest-to-predict games feature either repetitive sections of navigation with limited action (such as shooting, reloading, collecting items, or dodging fire). These games also tend to feature dark backgrounds and enemies with silhouettes that are difficult to distinguish or drab environments where the ground, background, and often non-player characters blend together.
\begin{figure}[t]
\centering
\includegraphics[width=1\linewidth]{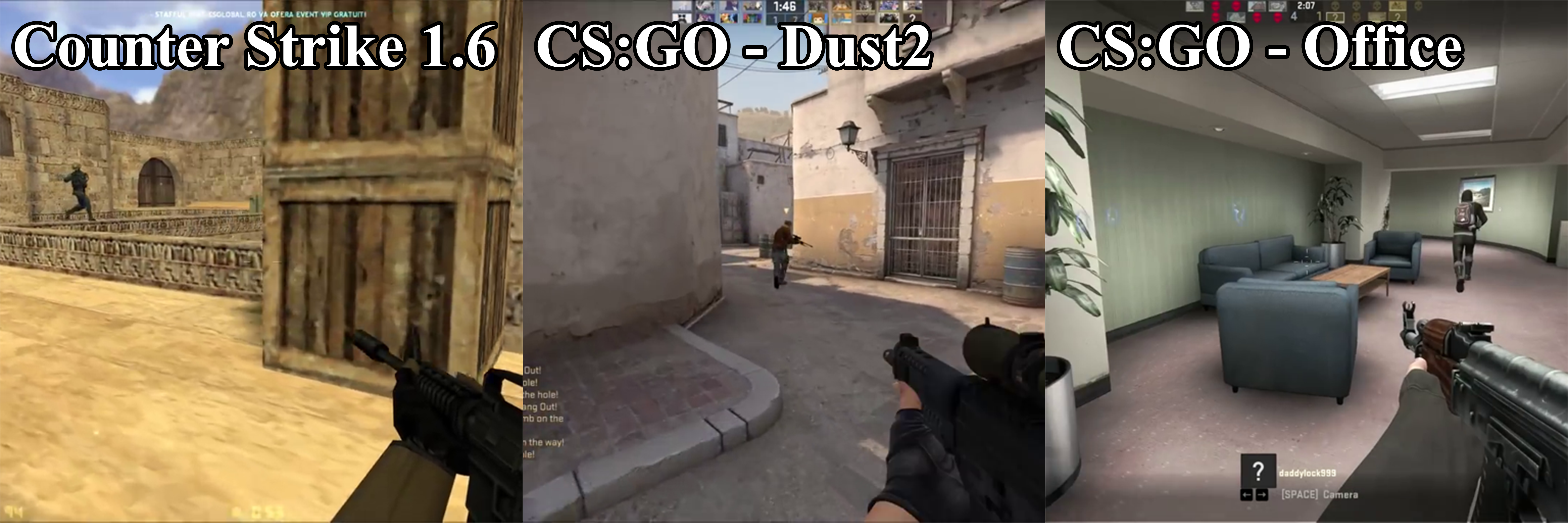}
\caption{Similar frames from \emph{Counter Strike }variants: \emph{Counter Strike 1.6}, \emph{CS:GO - Dust2}, and \emph{CS:GO - Office} (left to right)}\label{fig:counter_strike}
\end{figure}
A representative example that highlights these performance differences are the \emph{Counter Strike} game variants (see Fig.~\ref{fig:counter_strike}). Compared to \emph{Counter Strike 1.6} {($F1=0.51$) which is comparable to the \emph{DINOv2+MPL} and below the \emph{CNN Baseline}}, \textit{GPT-4o} surpasses the baseline models on \emph{CS:GO - Dust2} ($F1=0.56$). Even though these two games use essentially the same level, the visuals of \emph{CS:GO - Dust2} are much clearer, while in \emph{Counter Strike 1.6} the background and foreground are harder to separate.
In \emph{CS:GO - Office}---where the visuals are arguably even more readable---the model performs even better ({$F1=0.62$}).

Another way to explain the fluctuation in LLM performance is the familiarity of the model with the games per se. We observe that more popular games
(such as \emph{CS:GO}, \emph{Apex Legends}, and \emph{PUBG}, with a peak viewership\footnote{Peak viewership refers to the historically highest number of concurrent viewers watching a stream. It is indicative of the maximum audience size.} of $1,914,861$, $674,070$, and $597,663$, respectively on Twitch\footnote{\label{twitchtracker}Numbers retrieved from \url{https://twitchtracker.com/}, January 2025.}) are easier to predict compared to less popular games 
(such as \emph{HROT}, \emph{Heretic}, and \emph{Corridor 7}, with a peak viewership of $24,721$, $2,280$, and $195$ on Twitch\footref{twitchtracker}).
However, we should be careful with naive over-generalizations from these findings. For one, \emph{Counter Strike 1.6} is a variant of a very popular game with a peak viewership of $125,378$ in itself, but the models struggle with it. While it is possible that the training data of \emph{GPT-4o} contains images from more popular games, attempting to verify this by reconstructing parts of the \emph{GPT-4o} training data is out of the scope of this paper.

\section{Discussion}\label{sec:discussion}
The evaluation experiments presented in this paper are the first of its kind for LLM-based engagement prediction in games. While collectively we tried $4,840$ combinations of experimental settings---varying the model type, prompting strategy, input type, game, and ground truth---there are still many aspects that we did not explore in this initial study. We argue, however, that we set out to lay ground work for future research by approaching the problem of automating gameplay annotation in a relatively straightforward way. 

\subsection{Limitations}
The study presented here constitutes a preliminary look into how out-of-box LLMs can solve affective computing tasks relying on their knowledge priors. Naturally, the results of such a study face multiple limitations.

\subsubsection{Static Images}
We experimented with several models and prompting strategies but kept the granularity of the vision input constant. We focused on single frames sampled at a 3-second interval because the low resolution input is less costly and more readily available than high frames-per-second videos. Nevertheless, there is a high likelihood that some of the context of the action is lost during this processing. This is a core limitation of this initial study. Although we briefly experimented with different time intervals (i.e. between 1 and 5 seconds), simply increasing the sampling rate did not yield a performance increase. It is likely, however, that by either providing more frames per query or using video input directly would lead to a significant performance improvement that remains to be tested in future studies.

\subsubsection{No Fine-Tuning}
As our focus was to evaluate the state of LLMs relying solely on their own knowledge priors, we did not employ any form of fine-tuning. The main motivation for this decision is that we are interested in the zero-shot capabilities of available LLMs and their limits using prompt-engineering. While we can expect models with specialised fine-tuning to outperform general models, this is not always the case. Large commercial LLMs using few-shot prompting have shown greater flexibility and superior performance compared to pre-trained models that have been fine-tuned on specialised datasets in certain domains \cite{wang2023chatgpt}. As full-parameter fine-tuning is prohibitively costly, it is worthwhile investigating other avenues of making use of these models. Here we demonstrated prompt-engineering including a few-shot prompting strategy, but future studies should investigate efficient parameter tuning, such as low-rank adaptation \cite{muller2024recognizing, parthasarathy2024ultimate}.

\subsubsection{Limited Domain}
We showcase our results on $20$ different games with varying aesthetics and play-styles, however, the experiments focus only on the \emph{first-person shooter} genre. Even though this is a popular videogame genre---with countless examples out there, we show that positive results generalise poorly even within this limited domain. Our results show that clear visuals are key to the models' success. This holds true across different experiments and architectures. These results show the need for generalised game representation beyond simple pixels on the screen \cite{trivedi2021contrastive}.

\subsubsection{Baseline Performance}
Our choice of affect corpus, GameVibe, presents a challenging task for both the baseline and LLM models for several reasons. The most significant challenge is for the models to simultaneously generalize to both unseen video clips of the 20 games presented, and a new group of annotators due to the leave-one-session-out cross validation protocol used. Furthermore, the corpus is relatively small, making our smaller baseline models more susceptible to this distribution shift. These results demonstrate the strength of LLMs in challenging case studies where data is scarce, as all models outperformed the baselines tested. However, results also highlight that tests with larger corpora and more high-performing baselines are needed in the future.

With the exception of \textit{InternVL2.5}, which only significantly outperforms the \textit{Naive Baseline} under the \textit{Basic Description} text-based prompt condition, all evaluated LLMs significantly outperform the \textit{Naive Baseline}. This holds consistently across experimental conditions and prompt types, providing a reliable lower bound against which all models can be meaningfully compared. However, the picture is more nuanced when comparing against the stronger feature-based baselines. The \textit{CNN Baseline} and \textit{DINOv2+MLP} represent non-trivial competitors. Most LLMs perform statistically on par with these approaches under \textit{One-Shot} and \textit{Text Description} prompt conditions. Only under the \textit{Few-Shot} setting does \textit{GPT-4o} significantly outperform both \textit{CNN Baseline} and \textit{DINOv2+MLP}, and \textit{Gemma 3} outperforms \textit{DINOv2+MLP}. These findings suggests that, for the majority of models and conditions, the LLMs' advantage over learned visual feature extractors is limited. The gap between LLMs and the stronger baselines narrows or disappears entirely for weaker open-source models. This is most glaring with \textit{InternVL2.5}, which---despite being built to challenge large closed-source models \cite{chen2024expanding}---fails to significantly outperform any baseline beyond the \textit{Naive Baseline} under most conditions.

\subsubsection{Borderline Results}
LLMs struggle with engagement prediction in games \cite{wang2026vision}, because it is a subjective, hidden mental state that cannot be directly observed and does not reliably match what is on the screen. Visually intense moments are not always engaging, while subtle situations can be highly engaging. Successfully predicting engagement might require contextual and temporal reasoning about goals, challenge, and prior events, which models struggle with, often relying on shallow visual cues instead. Moreover, engagement varies across players, games, and time, and even humans do not fully agree on it, making accurate prediction inherently challenging.

Our per-game analysis suggests that models partially succeed by exploiting visual distinctiveness rather than reasoning about engagement. Games with highly consistent and recognisable visual styles (e.g. \textit{Wolfram}, \textit{Borderlands 3}, \textit{Apex Legends}) are systematically easier to predict, while visually generic or ambiguous titles (e.g. \textit{Insurgency}, \textit{Corridor 7}, \textit{HROT}) remain hard to predict across all conditions. The frame-by-frame setup compounds this, as individual frames cannot convey a larger temporal context. The absence of significant differences between prompt conditions at the aggregate level suggests that no universal prompt strategy can reliably elicit relevant reasoning; and gains from any particular prompting method appear highly game-dependent. Our experiments with a two-step procedure (first describe; then reason based on text descriptions) illustrate this: while this strategy yielded marginal gains in some instances, it reduced performance on games where visual cues are more informative. Finally, the variation in model performance (particularly \textit{InternVL2.5}'s failure to consistently outperform the \textit{Naive Baseline}) highlights a fundamental mismatch between standard multimodal benchmarks, which reward object recognition and spatial reasoning, over the affective, subjective and temporal nature of engagement prediction.

\subsection{Future Work}
While video input could feed more information to the LLM, the context of the query could also be augmented and then provided to the LLM, thereby improving its predictive capacity. By implementing a memory mechanism \cite{zhang2024survey}, we could potentially store and recall the temporal context of the play session. Similarly, we could provide more context on the necessary domain knowledge for the task by implementing \emph{retrieval-augmented generation} \cite{lewis2020retrieval}, where we could feed more information on the game or downstream task. We plan to pursue these avenues in our future studies in our effort to further investigate how more contextual information impacts the performance of LLMs towards fully autonomous engagement annotation.

Generating subjective labels is a relatively open field with a lot of unanswered questions. Naturally, the exploration should be extended to other datasets. While this study focusses on engagement, there are other subjective aspects of both player and viewer experience that could be evaluated further. A natural step forward would be to make use of a diverse set of affective corpora, focusing, for instance, on affect prediction across videogame datasets \cite{melhart2022again}, but also architectural spaces \cite{xylakis2024affect} and movie corpora \cite{girard2023dynamos}.
As discussed above, the current evaluation of LLMs---even though multimodal---considers a predetermined number of modalities: text and images. As we move forward and more multimodal architectures become widely adopted, the research into utilizing LLMs for autonomous affect annotation could encompass different modalities from images, through video, to audio. When it comes to interactive media such as games, user behavioural data could also be included \cite{ravsajski2024behave} providing a richer context to the models.

\section{Conclusions}\label{sec:conclusions}

This paper explored a novel application of LLMs for autonomously annotating the continuous experience of viewers when consuming videos of first-person shooter video games from the \emph{GameVibe} corpus. We conducted an in-depth analysis comparing multiple open-source foundation models---comparing them to Open-AI's \emph{GPT}---and evaluated their performance across different input modalities (i.e. multimodal, text-based) and prompting strategies (i.e. one-shot, few-shot). 
While the results obtained might be far from outstanding, we show that most models outperform traditional ML baseline methods. In fact, in our most successful experimental setup (few-shot), most models significantly outperform a naive majority-class baseline; \textit{Gemma 3} significantly outperforms a transformer-based classifier; and \textit{GPT-4o} significantly outperforms all baseline methods introduced in the paper.
This shows the potential of LLMs in otherwise complex affect modelling tasks.
Our findings confirm that beyond model size and prompting strategy, the breadth and diversity of training objectives also have a critical impact on model performance on subjective tasks, such as engagement prediction. 

The LLMs tested here demonstrate promising capabilities in certain game elicitors.
Perhaps unsurprisingly, the games where LLMs are successful in predicting the continuous change in engagement are popular games with easy-to-read graphical styles and concise gameplay. The gap in performance on these games compared to more challenging elicitors shows that while LLMs have potential, there is still a long road ahead towards automated continuous affect labelling using these type of foundation models.

As LLMs continue to scale and evolve, we believe their ability to capture subjective experiences will drastically improve---especially when incorporating richer multimodal inputs such as video, audio, and physiological signals. The annotation capacities of such LLM-based foundation affect models extends well beyond the domain of games to video-based general affect modelling, and human-computer interaction at large. Based on the results presented here, we believe future work could leverage LLMs as flexible and scalable annotators in a wide variety of dynamic and real-world settings.

\bibliographystyle{IEEEtran}
\bibliography{bibtex/paper}


%
\begin{IEEEbiography}[{\includegraphics[width=0.9in,clip,trim={40px 0 40px 0},keepaspectratio]{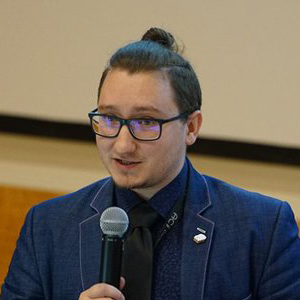}}]
{David Melhart} is a founding member and Senior AI Engineer at \emph{humanfeedback.ai} and Assistant Professor at the University of Southern Denmark (SDU). He received his Ph.D. from the University of Malta, focusing on user research and affective computing. He has contributed to various academic and industry events, as Communication Chair (\emph{FDG} 2020, 2022; \emph{DiGRA} 2025; \emph{IEEE CoG} 2026), Workshop and Panels Chair (\emph{FDG} 2023), and Workshop Organizer (\emph{The 1st ECAI Workshop on Artificial Intelligence for Human-Game Interaction} 2024; \emph{CHI-Play Workshop on Ethics and Transparency in Game Data} 2024). He is one of the main organizers and chief coordinator of the \emph{Summer School on AI and Games} (2018–2027). He serves as Editorial Assistant of \emph{IEEE Transactions on Games} and as an Associate Editor of \emph{IEEE Transactions on Affective Computing}.
\end{IEEEbiography}
\vskip -2\baselineskip plus -1fil
\begin{IEEEbiography}[{\includegraphics[width=0.9in,clip,trim={170px 0 170px 0},keepaspectratio]{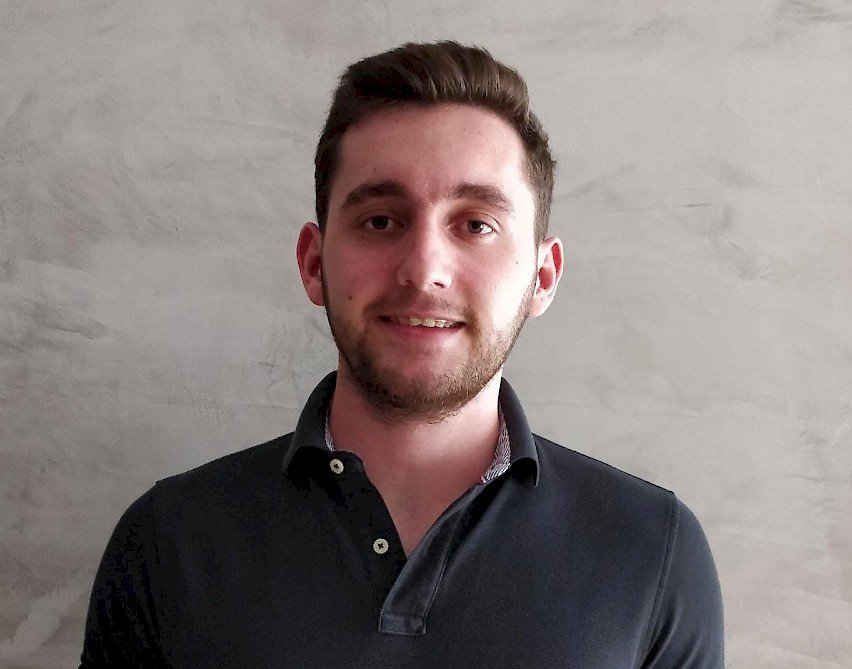}}]
{Matthew Barthet}
received a bachelors of science degree in computer science, and a masters of science degree in digital games from the University of Malta in 2019 and 2021, respectively. He is currently a PhD candidate at the University of Malta researching training reinforcement learning agents in affective computing applications. His other research interests include procedural content generation, game artificial intelligence, and computational creativity. 
\end{IEEEbiography}
\vskip -2\baselineskip plus -1fil
\begin{IEEEbiography}[{\includegraphics[width=0.9in,clip,trim={80px 60px 70px 20px},keepaspectratio]{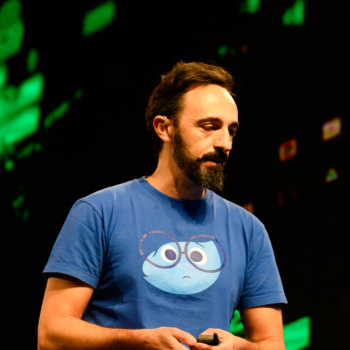}}]
{Georgios N. Yannakakis} is the Founding CEO of \emph{humanfeedback.ai} and Professor at the Institute of Digital Games, University of Malta (UM). He received the PhD degree in Informatics from the University of Edinburgh in 2006. Prior to joining UM, in 2012 he was an Associate Professor at the Center for Computer Games Research at the IT University of Copenhagen. He does research at the crossroads of artificial intelligence, affective computing, games and computational creativity. He has published more than 300 papers in the aforementioned fields and his work has been cited broadly. His research has been supported by numerous national and European grants (including a Marie Skłodowska-Curie Fellowship) and has appeared in \emph{Science Magazine} and \emph{New Scientist} among other venues. He was the Editor in Chief of the \emph{IEEE Transactions on Games} (2022-2025), and an Associate Editor of the \emph{IEEE Transactions on Evolutionary Computation}, the \emph{IEEE Transactions on Affective Computing} and the \emph{IEEE Transactions on Computational Intelligence and AI in Games} journals. 
Among the several rewards he has received for his papers he is the recipient of the \emph{IEEE Transactions on Affective Computing Most Influential Paper Award} and the \emph{IEEE Transactions on Games Outstanding Paper Award}. Georgios is an IEEE Fellow.
\end{IEEEbiography}





\end{document}


%
\title{Can Large Language Models Capture \\Video Game Engagement? - Appendixes}

%
%
%


\author{
    \IEEEauthorblockN{
        David Melhart\IEEEauthorrefmark{1},
        Matthew Barthet\IEEEauthorrefmark{2},
        and Georgios N. Yannakakis\IEEEauthorrefmark{1} \emph{IEEE Fellow}
    }
    \\
    \IEEEauthorblockA{\IEEEauthorrefmark{1}
        \emph{humanfeedback.ai}, \IEEEauthorrefmark{2}
        \emph{Institute of Digital Games, University of Malta}\\
        Malta\\
        david@humanfeedback.ai, matthew.barthet@um.edu.mt, georgios@humanfeedback.ai
    }
}
\setcounter{figure}{9}

\maketitle

\appendix  

\begin{figure}[ht]
\centering
\includegraphics[width=0.8\linewidth]{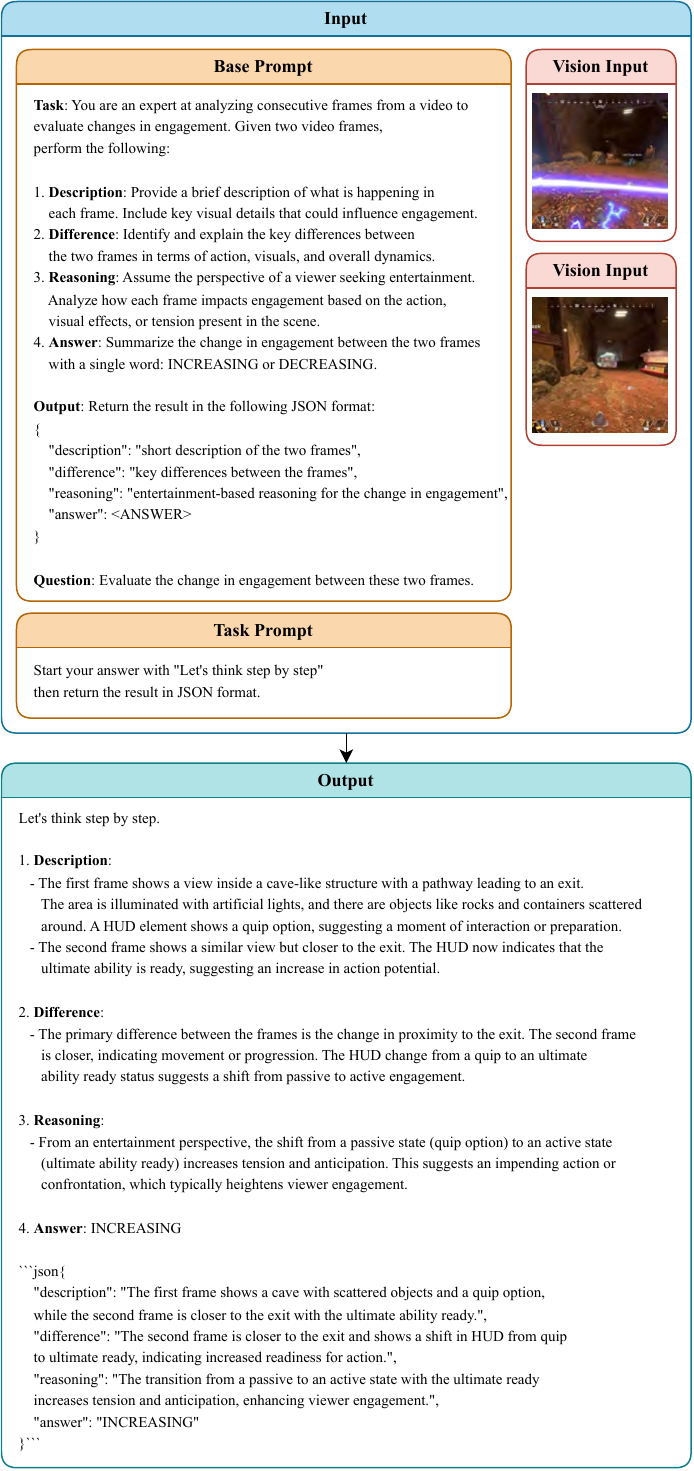}
\caption{Multimodal One-Shot input and output example using \emph{GPT-4o}.}\label{annex:oneshot}
\end{figure}

\begin{figure}[ht]
\centering
\includegraphics[width=0.8\linewidth]{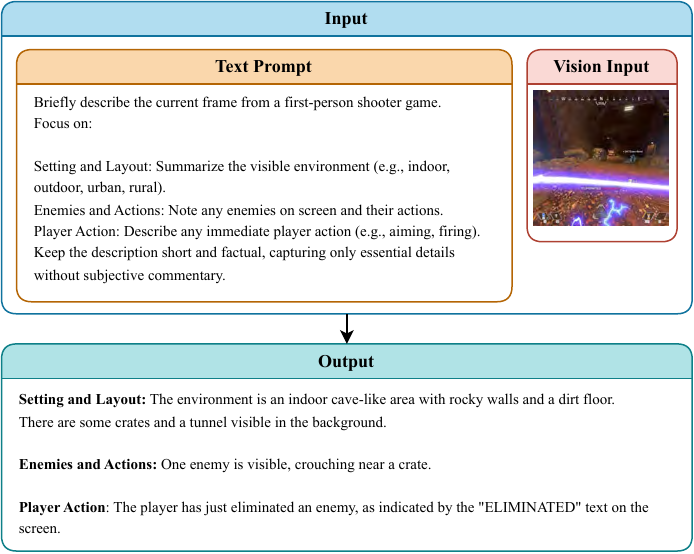}
\caption{Basic Text Description image-to-text description generation using \emph{GPT-4o}.}\label{annex:text-basic}
\end{figure}

\begin{figure}[ht]
\centering
\includegraphics[width=0.8\linewidth]{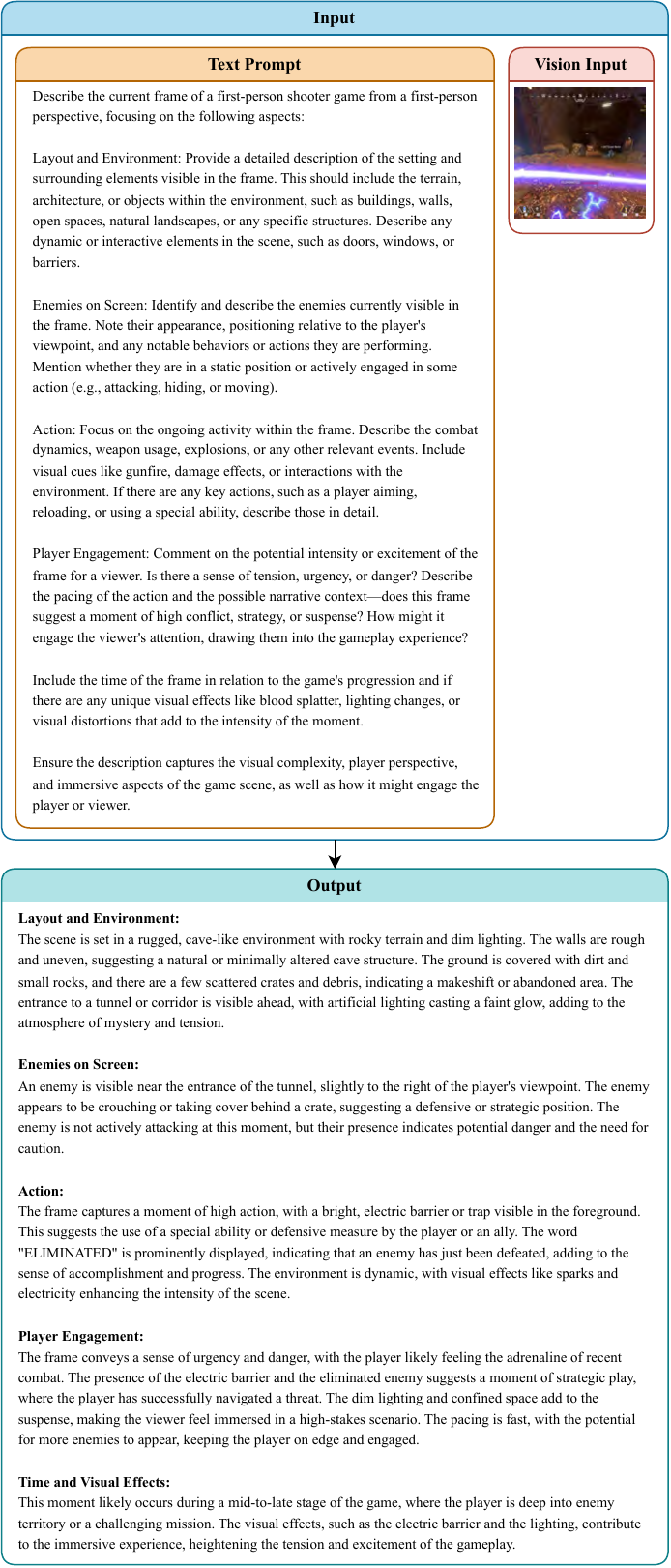}
\caption{Advanced Text Description image-to-text description generation using \emph{GPT-4o}.}\label{annex:text-advanced}
\end{figure}

\begin{figure}[ht]
\centering
\includegraphics[width=0.8\linewidth]{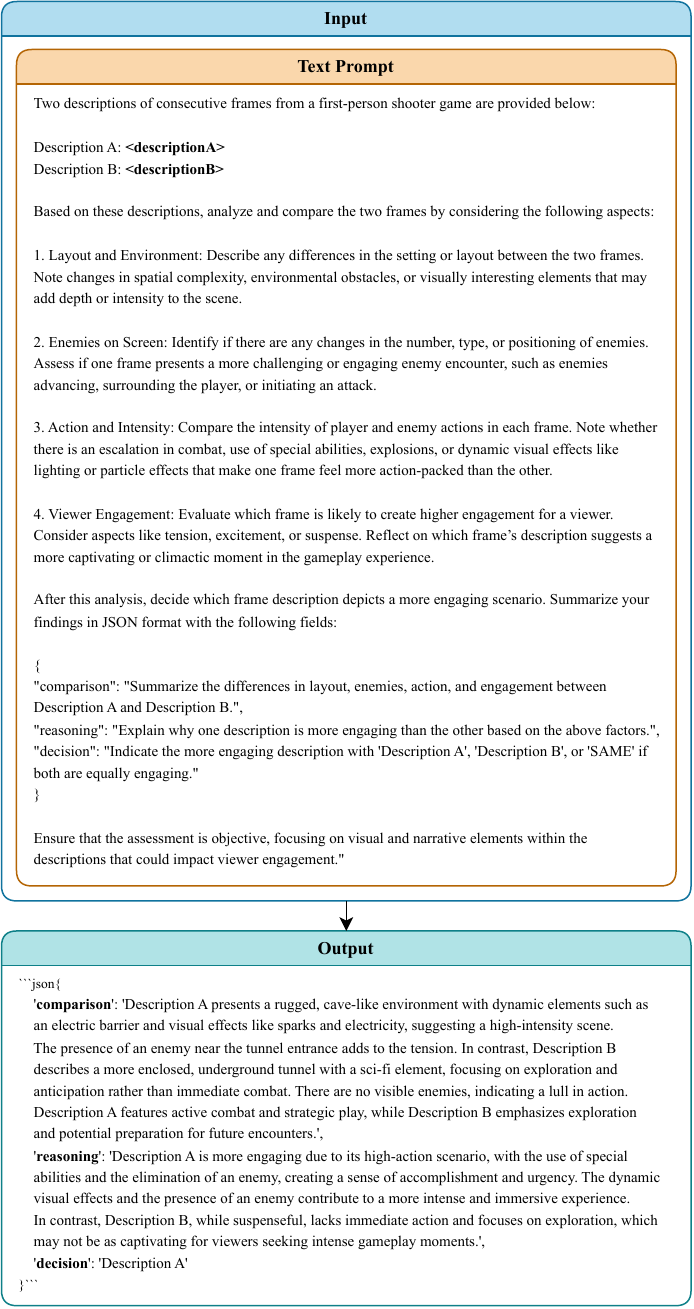}
\caption{Text Input - Engagement Prediction using \emph{GPT-4o}. $<$description A$>$ and $<$description B$>$ are replaced with generated text descriptions.}\label{annex:text-prediction}
\end{figure}

\begin{figure}[ht]
\centering
\includegraphics[width=0.8\linewidth]{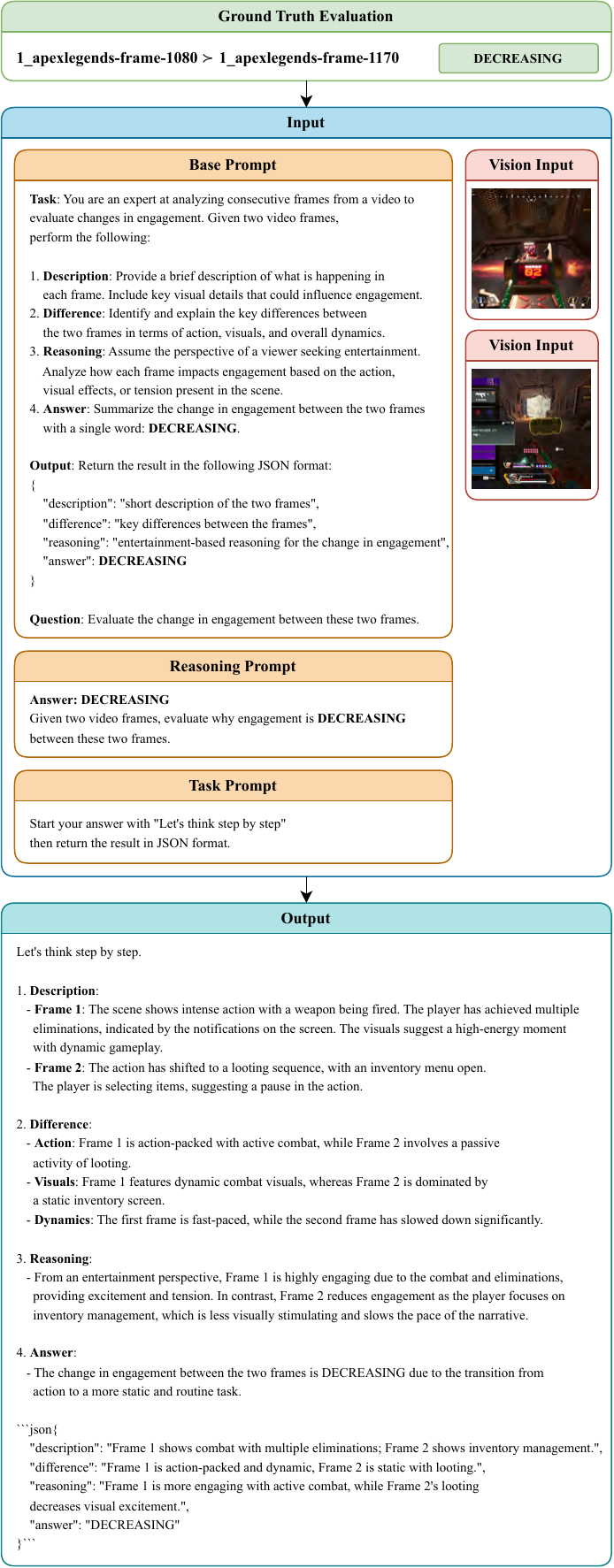}
\caption{Multimodal Few-Shot Example Reasoning Generation input and output example using \emph{GPT-4o}. The generated output is used as an artificial example in the Few-Shot experiments.}\label{annex:fewshot-examples}
\end{figure}

\begin{figure}[ht]
\centering
\includegraphics[width=0.8\linewidth]{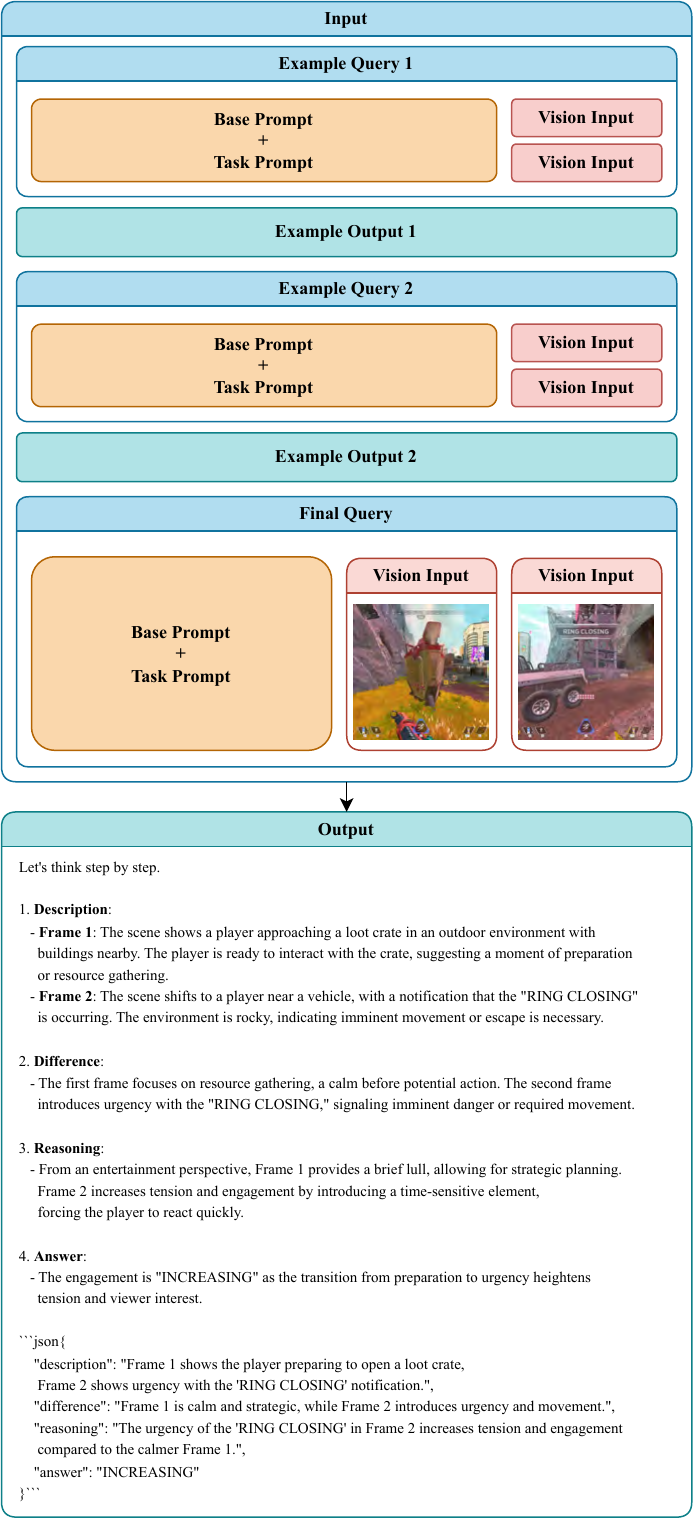}
\caption{Multimodal Few-Shot Engagement Prediction input and output example using \emph{GPT-4o}. The Base and Task Prompts are identical to the ones described in Figure~\ref{annex:oneshot}. Example Outputs are generated as shown in Figure~\ref{annex:fewshot-examples}.}\label{annex:fewshot-prediction}
\end{figure}